\pdfoutput=1

\documentclass[11pt]{article}

\usepackage{acl}

\usepackage{amsmath}
\usepackage{amssymb}

\usepackage{times}
\usepackage{latexsym}

\usepackage[T1]{fontenc}

\usepackage[utf8]{inputenc}

\usepackage{microtype}

\usepackage{inconsolata}

\usepackage{graphicx}
\usepackage{booktabs}
\usepackage{multirow}
\usepackage{tabularx}
\usepackage{svg}
\usepackage{threeparttable}

\usepackage{subcaption}
\title{Uncovering the Role of Initial Saliency in U-Shaped Attention Bias: Scaling Initial Token Weight for Enhanced Long-Text Processing}

\author{
  Zewen Qiang,
  Sendong Zhao\thanks{Corresponding author.},
  Haochun Wang,
  Bing Qin and
  Ting Liu \\
  \texttt{zwqiang@ir.hit.edu.cn}
}

\begin{document}
\maketitle
\begin{abstract}
Large language models (LLMs) have demonstrated strong performance on a variety of natural language processing (NLP) tasks. However, they often struggle with long-text sequences due to the ``lost in the middle'' phenomenon. This issue has been shown to arise from a U-shaped attention bias, where attention is disproportionately focused on the beginning and end of a text, leaving the middle section underrepresented. While previous studies have attributed this bias to position encoding, our research first identifies an additional factor: initial saliency. It means that in the attention computation for each token, tokens with higher attention weights relative to the initial token tend to receive more attention in the prediction of the next token. We further find that utilizing this property by scaling attention weight between the initial token and others improves the model's ability to process long contexts, achieving a maximum improvement of 3.6\% in MDQA dataset. Moreover, combining this approach with existing methods to reduce position encoding bias further enhances performance, achieving a maximum improvement of 3.4\% in KV-Retrieval tasks.
\end{abstract}

\section{Introduction}
Large language models (LLMs)~\cite{Radford_Narasimhan_Salimans_Sutskever,achiam2023gpt,llama,team2024gemini} excel in diverse NLP tasks, including tool usage~\cite{bucket}, extended dialogues~\cite{Zhong_Liu_Xu_Zhu_Zeng_2022}, and domain-specific applications like healthcare~\cite{wang2024knowledge}. However, many tasks require integrating external information to update outdated data or inject specialized knowledge. Retrieval-augmented generation (RAG) enhances LLMs by incorporating relevant documents, significantly improving performance~\cite{gao2023retrieval,fan2024survey}.
\begin{figure}[t]
    \centering
    {\includegraphics[width = 1.\linewidth]{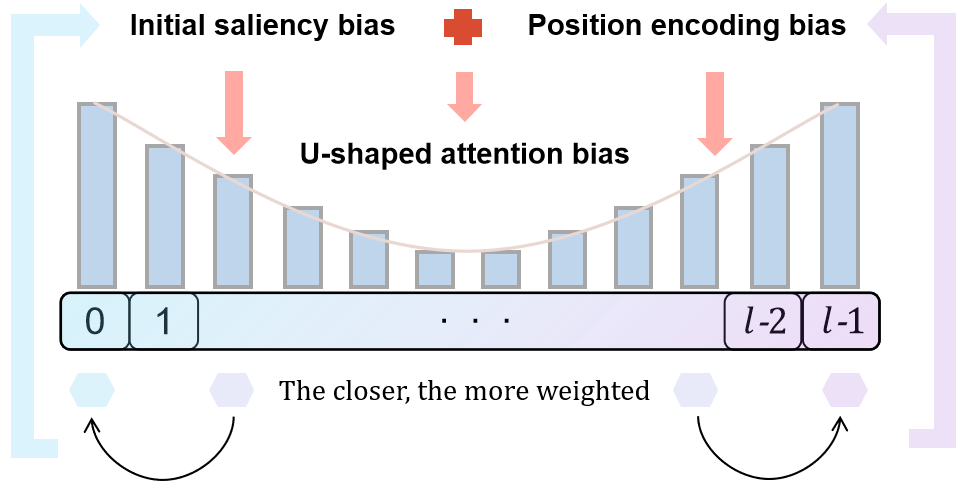}}
    \caption{Aside from the widely discussed position encoding bias, initial saliency is also a significant factor contributing to the U-shaped attention bias. Position encoding bias causes tokens closer to the last token to receive higher attention weights. In contrast, initial saliency bias leads to tokens nearer to the initial token receiving more attention. These two biases result in U-shaped attention waves together.}
    \label{intro}
\end{figure}
\begin{figure*}[htb]
    \centering
    \subcaptionbox{Llama-7B\label{bar_line_1}}{\includegraphics[width = .241\linewidth]{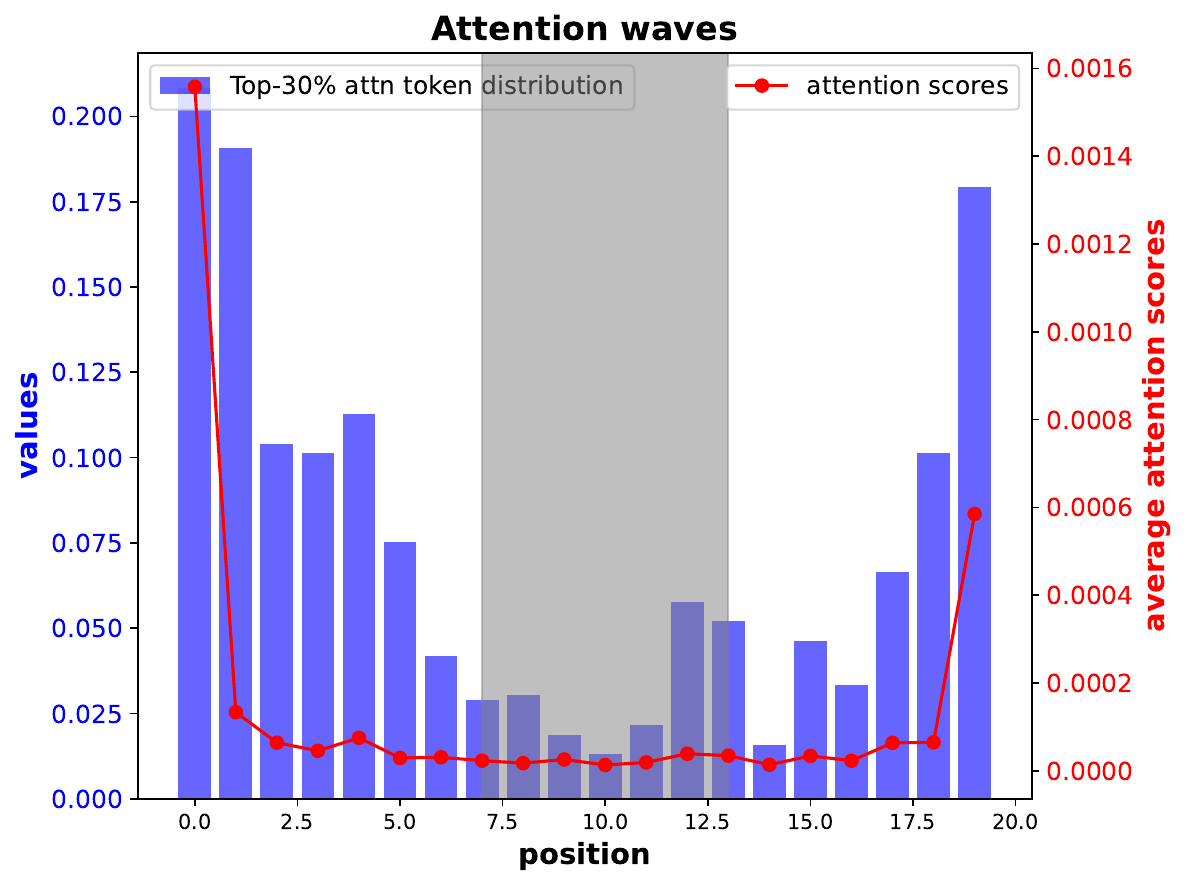}}
    \hspace{0.0001mm}
    \subcaptionbox{Vicuna-7B\label{bar_line_2}}{\includegraphics[width = .241\linewidth]{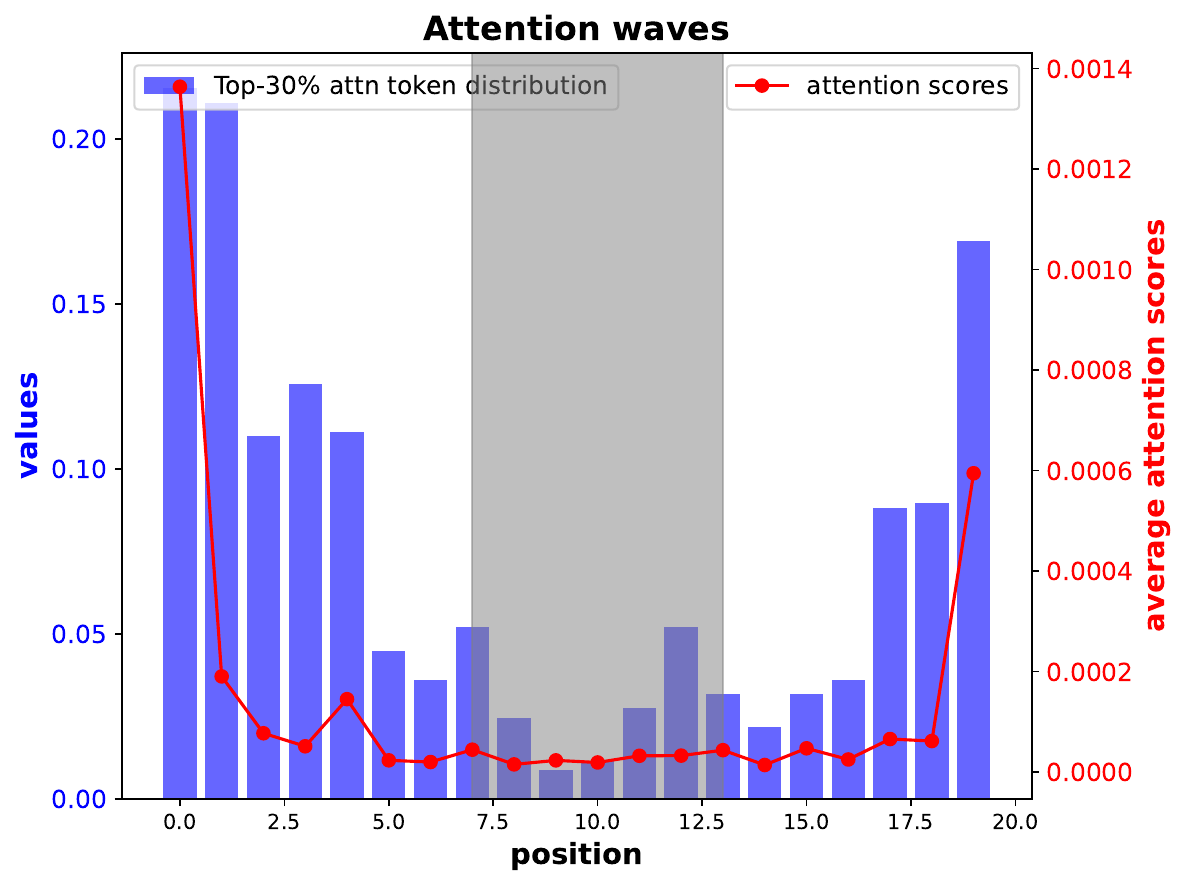}}
    \hspace{0.0001mm}
    \subcaptionbox{Tulu-7B\label{bar_line_3}}
    {\includegraphics[width = .241\linewidth]{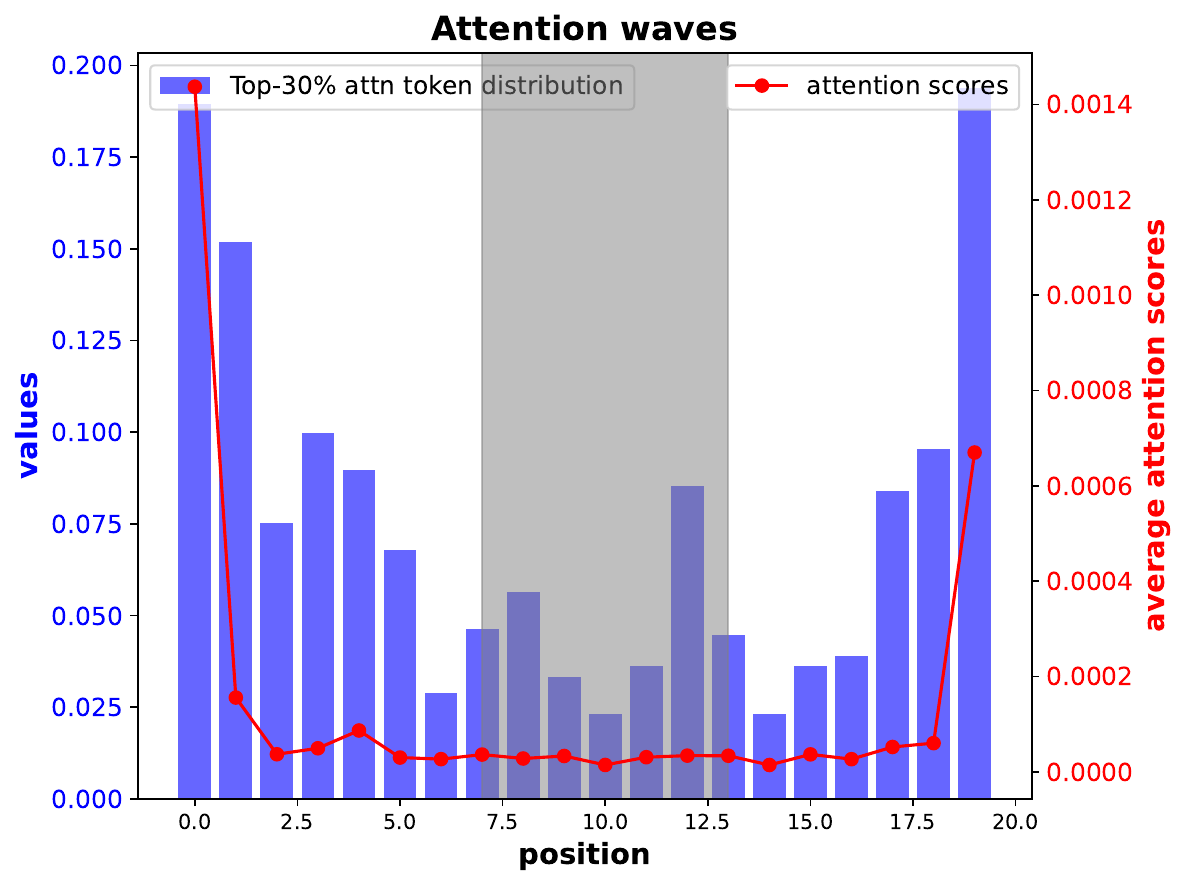}}
    \hspace{0.0001mm}
    \subcaptionbox{Human Serial Position Effect In Free Recall\label{bar_line_4}}
    {\includegraphics[width = .241\linewidth]{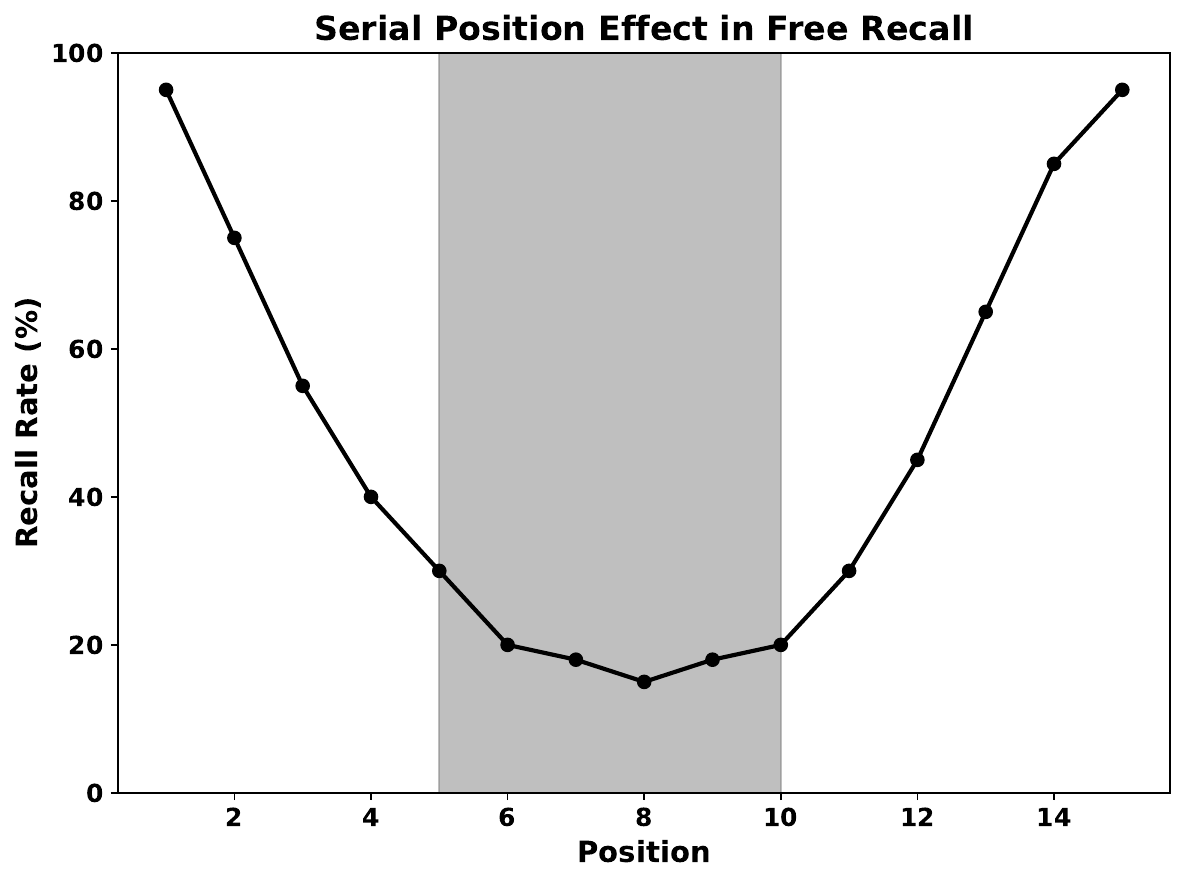}}
    \caption{ The attention curves for Llama-7B, Vicuna-7B, and Tulu-7B on the MDQA dataset are presented, along with the recall rates at different positions in Human-free recall tasks. The line chart represents the average attention weights of tokens across different documents, while the bar chart shows the distribution of the top 30\% attention-weighted tokens at various positions. It can be observed that both exhibit the same U-shaped pattern.}
    \label{fig0}
\end{figure*}
Despite these advancements, recent studies~\cite{lostinthemiddle,shi2023large,li-etal-2024-loogle,li2024longcontext} have shown that LLMs struggle to effectively utilize long-text information, particularly from intermediate sections—a phenomenon commonly referred to as ``lost in the middle.'' This issue has been linked to the U-shaped attention bias~\cite{middle}. It means the inherent U-shaped attention distribution observed in LLMs, akin to the primacy and recency effects~\cite{asch1946forming,baddeley1993recency} in human cognition. Humans tend to focus attention on the first and last impressions, with reduced attention to the intermediate content. Similarly, LLMs allocate lower attention weights to tokens in the middle of a sequence. 

The U-shaped attention curve is often attributed to position encoding bias~\cite{bucket}. This hypothesis has been validated by several studies~\cite{se,mspoe}, which have shown that tokens farther from the last token in relative position tend to receive less attention. This leads to the concentration of attention on tokens at the end of the sequence. However, they fail to fully explain why tokens at the beginning of prompts consistently receive higher attention weights or why the model is more efficient at leveraging initial content.

In this work, we identify that the U-shaped attention curve is not solely caused by position encoding bias but also strongly influenced by the \textit{initial saliency}. It means that in the attention computation for each token, tokens with higher attention weights relative to the initial token tend to receive more attention in the prediction of the next token. This property causes attention to concentrate on the earlier parts of the prompt. Since in decoder-only models, the causal mask causes tokens closer to the initial token to receive higher attention with initial token in their own attention computations. The uneven allocation of attention weights between the initial token and other tokens causes the initial saliency to dominate in the attention computation, thereby exacerbating the U-shaped attention bias.

 Additionally, humans consciously balance their attention across different stages of a task to overcome the primacy and recency effects. Similarly, we explore whether scaling the attention weights between the initial token and other tokens can improve the model's ability to leverage middle-content information. Experiments across different models show a 3.6\% improvement in MDQA~ \cite{lostinthemiddle}at most. 

All in all, our contributions are threefold: 
\begin{itemize}
    \item We introduce the concept of initial saliency and provide a more comprehensive explanation for the U-shaped attention curve in LLMs, showing that both initial saliency and position encoding bias contribute to the U-shaped attention bias.
    \item We first find that scaling the attention weights associated with the initial token can calibrate attention and mitigate the model's primacy and recency effects.
    \item We show that combining position information scaling and initial token weight scaling more effectively calibrates attention and enhances long-text modeling.
\end{itemize}
\begin{figure*}[htb]
    \centering
    \subcaptionbox{Llama-7B\label{2line_1}}{\includegraphics[width = .3\linewidth]{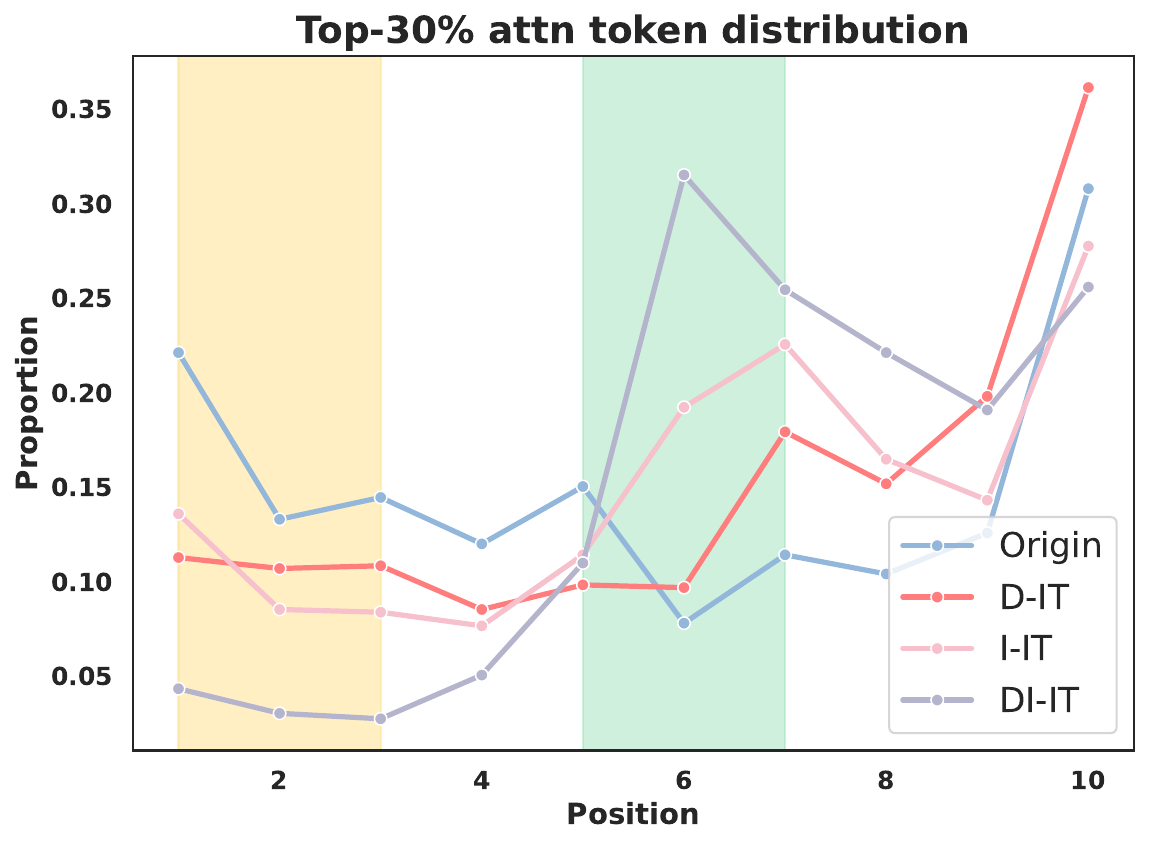}}
    \hspace{0.0001mm}
    \subcaptionbox{Vicuna-7B\label{2line_2}}{\includegraphics[width = .3\linewidth]{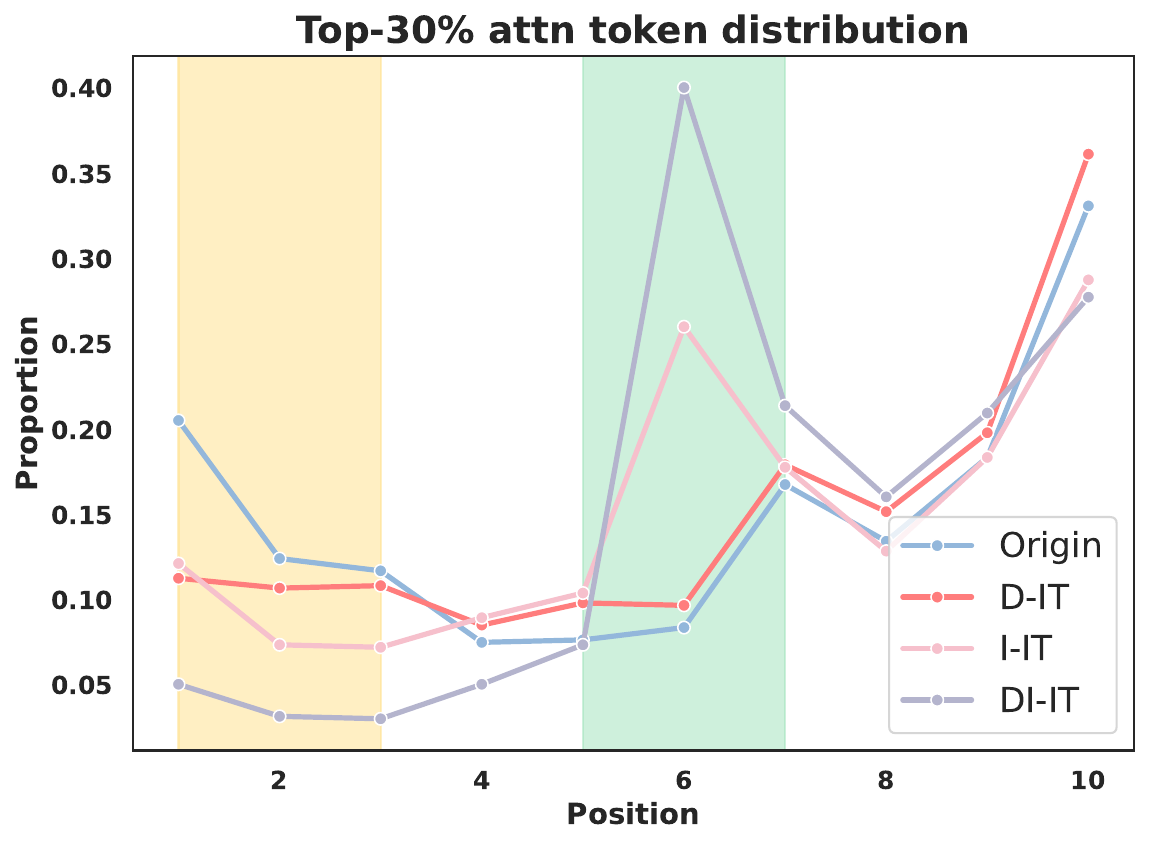}}
    \hspace{0.0001mm}
    \subcaptionbox{Tulu-7B\label{2line_3}}
    {\includegraphics[width = .3\linewidth]{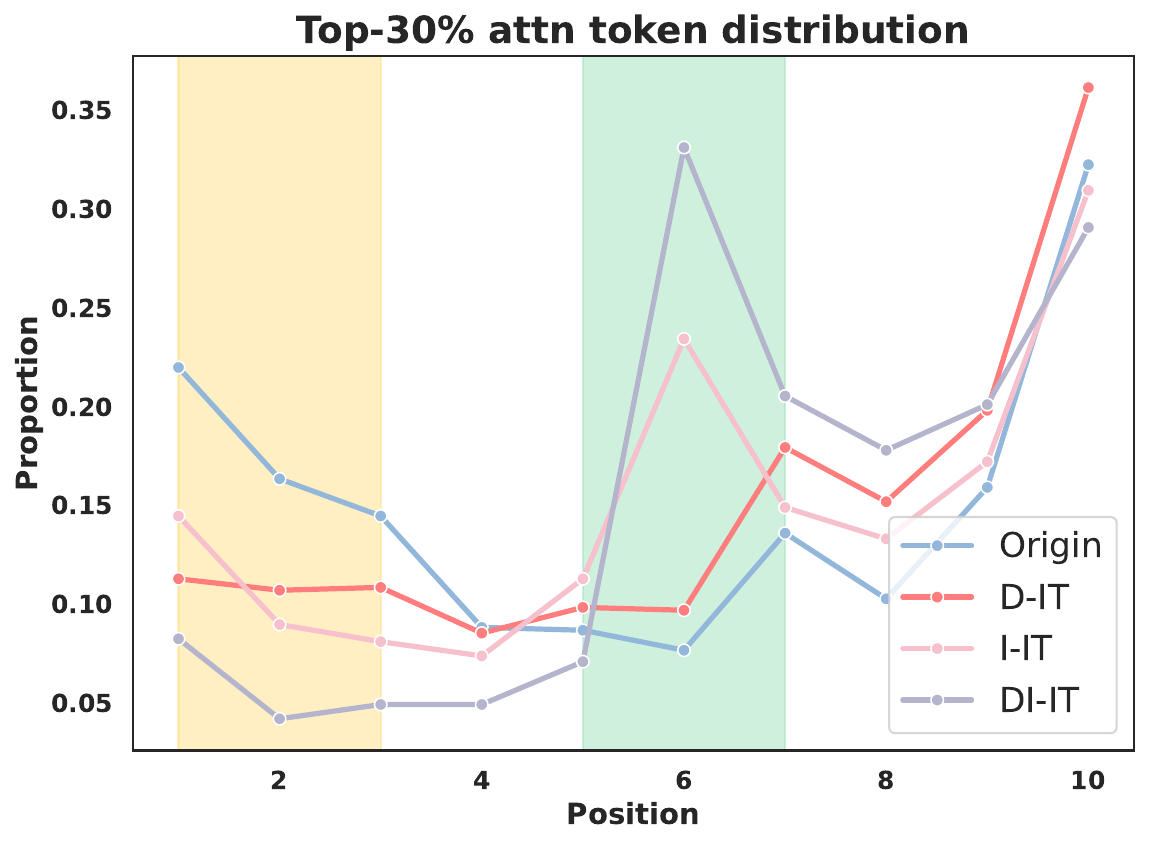}}
    \caption{ The distribution of attention enrichment points after scaling the attention weight between initial token and other tokens. As shown in the figure, increasing the attention weight between a token and the initial token allows the token to receive more attention during subsequent generation. Conversely, decreasing the attention weight between the token and the initial token in the opposite effect.}
    \label{fig2}
\end{figure*}
\section{Initial Saliency}
\label{AS}
This section primarily uses experiments to demonstrate the existence of initial saliency and analyze its impact on the U-shaped attention bias.
\subsection{Impact of Initial Saliency on Attention Distribution}
\textit{Initial saliency} refers to the phenomenon where, in the attention computation for each token, tokens with higher attention weights relative to the initial token are more likely to receive increased attention in the prediction of the next token. To demonstrate the existence of this phenomenon, we compared the attention curves for next token prediction after reducing and increasing the attention between different tokens and the initial token.

The experimental data consists of 200 question-answer pairs, each composed of 10 documents from the MDQA~\cite{lostinthemiddle} multi-document question-answering dataset. Experiment settings:(1) Origin, which means the base attention without any modification. (2) D-IT, which means decreasing the attention weights between the initial token and tokens in the first two documents. (3) I-IT, which means decreasing the attention weights between initial token and tokens in the 5th and 6th documents. (4) DI-IT, which means performing settings D-IT and I-IT simultaneously. This is evident from the global attention distribution in Figure \ref{fig0}, where the regions corresponding to the first two documents show a concentration of attention, while the regions for the 5th and 6th documents exhibit sparse attention. By applying reduction and enhancement operations in these two regions, the changes in the experimental results become more noticeable.

From the experimental results in Figure \ref{fig2}, we observe that reducing the attention weights between the initial token and tokens at the beginning of the sequence decreases their attention weights in the next token prediction. Conversely, increasing the attention weights between the initial token and tokens in the middle of the sequence enhances their attention weights in the next token prediction. When both operations are applied simultaneously, the changes in attention become more pronounced. The results provides a direct evidence for the existence of initial saliency, demonstrating that higher attention weights relative to the initial token increase a token's attention weight in the next token prediction, while lower attention weights reduce it.
\subsection{Primacy Effect and Recency Effect in LLMs: U-shaped Attention Bias}
Previous research~\cite{middle} has demonstrated that attention wave patterns are pervasive across various LLMs and often exhibit a U-shaped structure as shown in Figure \ref {fig0}. We test the first 200 examples from the MDQA~\cite{lostinthemiddle} multi-document question-answering dataset, where key information is placed in the middle positions, and presented the attention curves of different models~(more results in the Appendix~\ref{sec:barline}). Previous research has focused on the average attention of tokens within documents. However, as shown in the Figure \ref {fig0}, the average attention across large regions in the middle is not sufficiently distinct, making it difficult to capture fine-grained attention contrasts between different positions. To address this issue, we also show the proportion of tokens within each document that rank in the top 30\% of attention weights across the entire prompt. This is based on existing studies~\cite{snapkv} that demonstrate the greatest impact on model outputs comes from the distribution of high-attention tokens.
\subsection{Initial Saliency Also Contributes to the U-shaped Attention Bias}
Previous work~\cite{bucket,se} has demonstrated the impact of position encoding bias on the U-shaped attention bias. Tokens closer to the last token are more likely to receive higher attention in the next token prediction, which contributes to the concentration of attention in the latter half of the prompt.

The results above indicate that in addition to position embedding, initial saliency is another key factor that influences U-shaped attention bias. The reason why tokens in earlier positions tend to receive more attention is also easily explained. Formally, for an input prompt $x^{l}\in \mathbb{R}^{l\times d}$ whose length is $l$, the attention weight between the $i$-th token and the $l$-th token is denoted as $A_{i}^{l}\in \mathbb{R}$. In auto-regressive models, each token can only attend to the tokens preceding it, and the sum of the attention weights is always 1. 
\begin{equation}
   \sum_{i=0}^{l} A_{i}^{l} = 1
\end{equation}
Each token is assigned an attention weight greater than $\varepsilon(\varepsilon > 0)$. As shown in Equation \ref{1-attn}: As the input text length increases and $l$ becomes larger, $A_{0}^{l}$ tends to decrease. As a result, tokens farther from the initial token tend to allocate less attention weight to the initial token. On the other hand, tokens closer to the initial token exhibit the opposite trend.
\begin{equation}
\begin{aligned}
   A_{0}^{l} = 1 - \sum_{i=1}^{l} A_{i}^{l} \geqslant 1 - (l-1)\cdot \varepsilon 
   \label{1-attn}
\end{aligned}
\end{equation}
Therefore, tokens closer to the beginning of the sequence have relatively higher attention weights with respect to the initial token. This, combined with initial saliency, leads to a concentration of attention in the first half of the sequence during next token prediction.
\begin{figure}[t]
    \centering
    \subcaptionbox{Layer 1 attn-scores\label{1}}{\includegraphics[width = .48\linewidth]{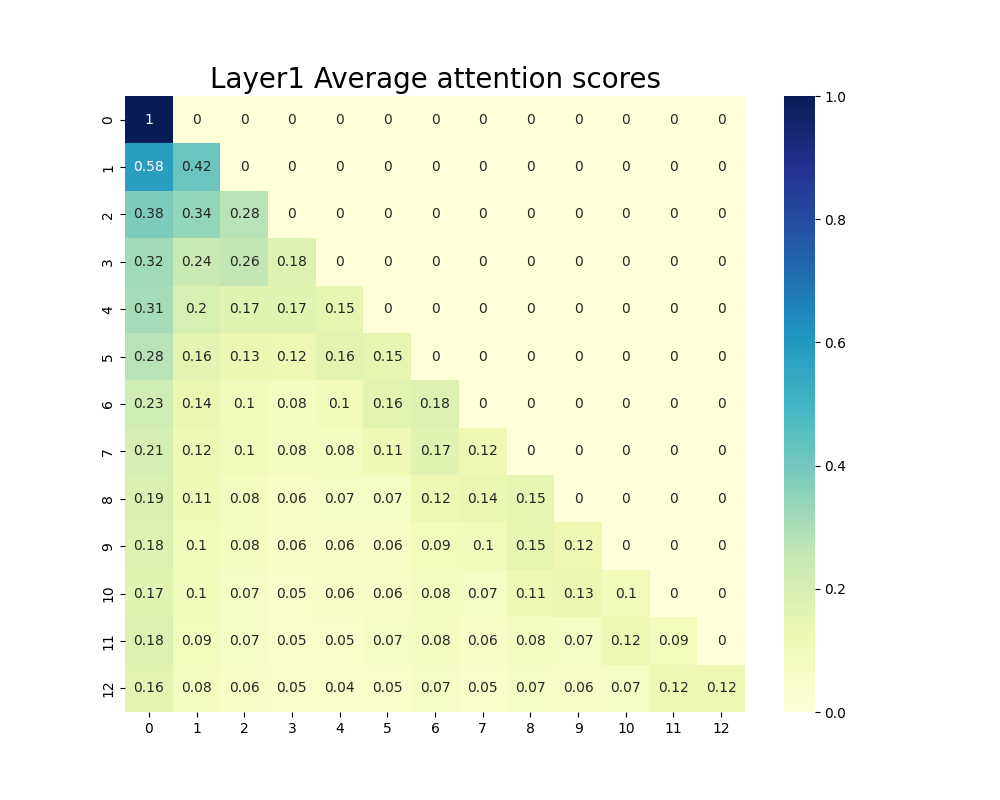}}
    \hspace{0.0001mm}
    \subcaptionbox{Layer 10 attn-scores\label{2}}{\includegraphics[width = .48\linewidth]{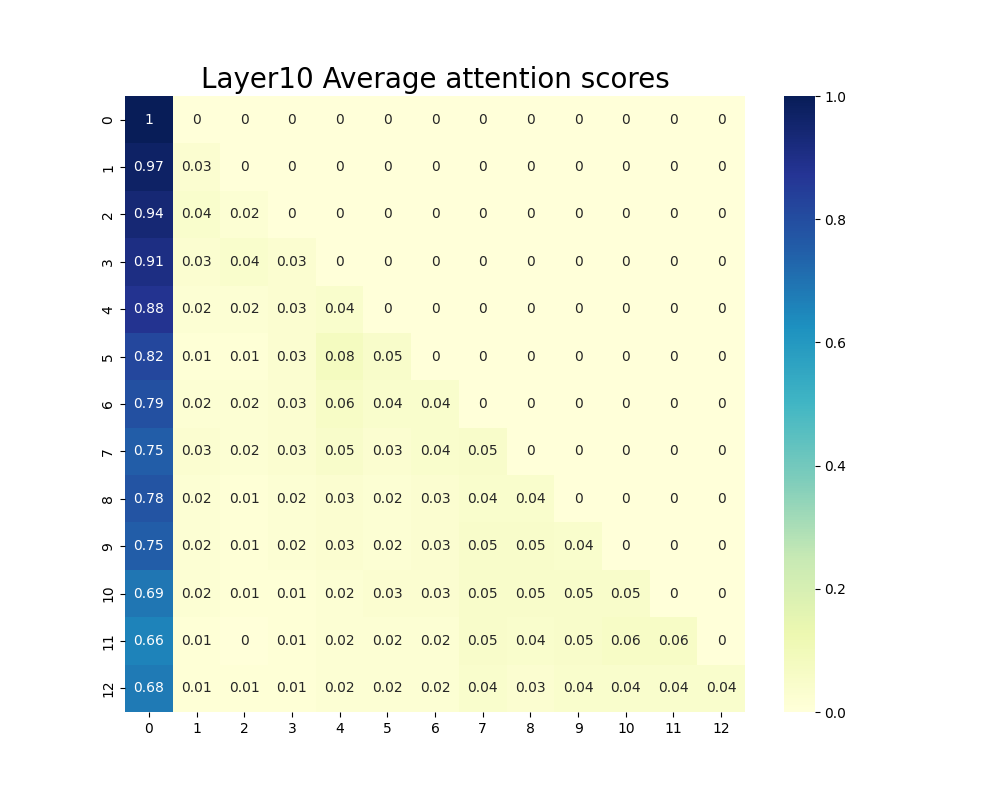}}
    \hspace{0.0001mm}
    \subcaptionbox{Layer 20 attn-scores\label{3}}
    {\includegraphics[width = .48\linewidth]{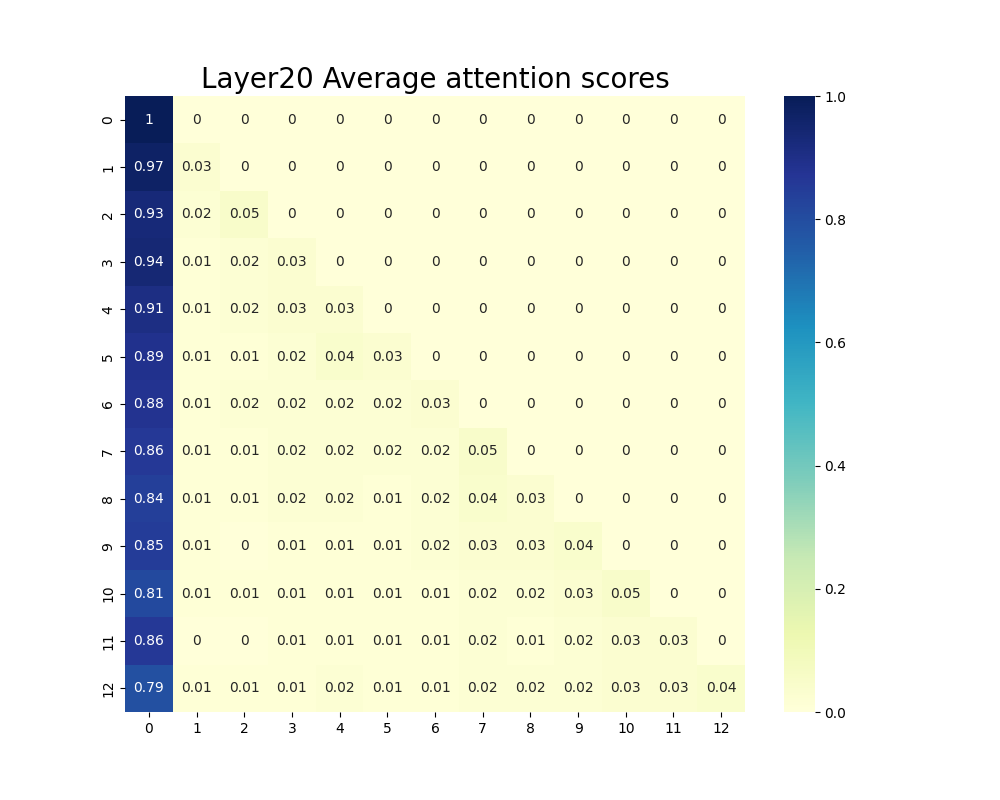}}
    \hspace{0.0001mm}
    \subcaptionbox{Layer 31 attn-scores\label{4}}
    {\includegraphics[width = .48\linewidth]{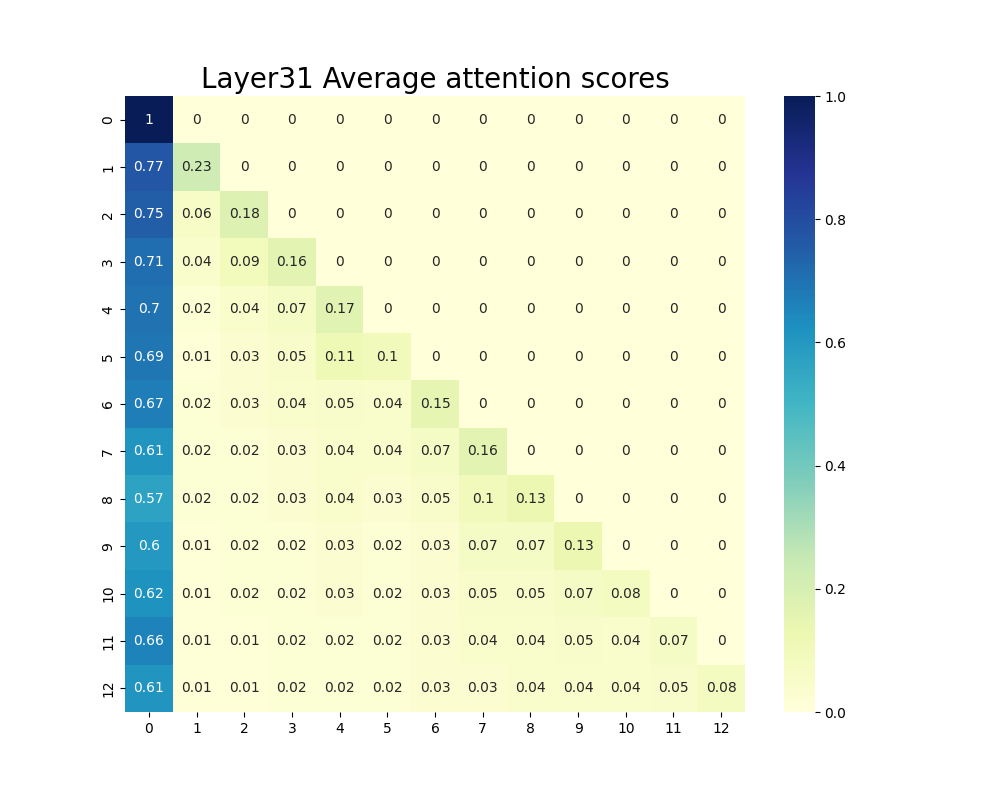}}
    \caption{ Visualization of the average attention in Llama2-7B-chat over 200 random sentences whose length are 13. The figure shows that, beginning with layer 1, each layer allocates considerable attention weights to the initial token.}
    \label{AS_f}
\end{figure}
\subsection{Causes of Initial Saliency Formation}
One possible cause of initial saliency formation is that the initial token often acts as an attention sink~\cite{scale_AS}. An attention sink~\cite{streamllm} refers to tokens that, despite lacking substantial semantic information, consistently receive high attention (Figure~\ref{AS_f}). These tokens are inherently more likely to attract attention, and as a result, the higher the attention weight between the token and the initial token during the token update process, the more likely it is that the token benefits from this ``attention favor''. This contributes to the formation of initial saliency.
\section{Investigation Into Attention Balancing with Scaling Initial Token Weight}
Humans often use attention balancing to counteract the primacy and recency effects. In this study, we explore the use of scaling the attention weights between tokens and the initial token to balance the model's attention, thereby mitigating the ``lost in the middle'' phenomenon. We refer to this process as \textbf{SIW} (Initial Token Weight Scaling).
\subsection{Problem Formulation}
Building on the previous Section~\ref{AS}, we can now provide a more comprehensive explanation of the underlying causes of this U-shaped attention bias. The factors influencing the calculation of attention can be represented by Equation \ref{oth}~\cite{middle}.
\begin{equation}
A_{i}^{l} =f(tok_{i}, i) = rel(tok_{i})+bias(i)+\tau
\label{oth}
\end{equation}
where function $f(\cdot)$ represents the attention computation between a specific token and the last token. The term $rel(\cdot)$ describes the relevance between a given token and the last token. $tok_{i}$ denotes the $i$-th token in the input sequence, while $bias(i)$ captures the impact of attention bias at position $i$.
Based on our analysis of attention bias, we can derive the following Equation \ref{two}. $\tau$ represents other residual terms.
\begin{equation} 
  bias(i)=pos(i)+sal(i)
  \label{two}
\end{equation}
where $pos(i)$ represents the influence of position encoding bias on the attention weight at position $i$, while $sal(i)$ denotes the impact of initial saliency. Both $pos(i)$ and $sal(i)$ exhibit monotonic behavior, with $pos(i)$ being monotonically increasing and $sal(i)$ monotonically decreasing. The opposing monotonic trends of these two factors combine to produce a U-shaped attention pattern.
\begin{equation} 
\left\{\begin{matrix} 
  pos(\mu\cdot i)+sal( i) <bias(i),\mu\in (0,1)\\  
  \varphi \cdot pos( i)+sal( i) <bias(i),\varphi\in (0,1)
\end{matrix}\right.  
  \label{mul}
\end{equation}

Previous work on mitigating attention bias can be summarized into two approaches, as outlined in Equation \ref{mul}. The first approach involves scaling the relative positions between tokens~\cite{mspoe,se}, thereby reducing the attention bias by leveraging the monotonicity of $pos(i)$. The second approach directly scales $pos(i)$ to reduce attention bias~\cite{caulsual_mask}. However, both approaches overlook $sal(i)$. Unlike $pos(i)$, $sal(i)$ is related to the absolute position of tokens and cannot be directly scaled by absolute position $i$. Therefore, our focus is twofold: (1) we demonstrate that directly scaling $sal(i)$ can effectively reduce attention bias; (2) our experiments show that combining both approaches yields improved results. This means that for \( \gamma \in (0,1) \), either \( \gamma \cdot sal(i) + pos(\mu \cdot i) \) or \( \gamma \cdot sal(i) + \varphi \cdot pos(i) \) performs better.

\subsection{Attention Balancing with SIW}
As previous research suggests~\cite{caulsual_mask,mspoe,middle}, a good attention balancing can significantly improve the ``lost in the middle'' problem. We attempt to mitigate the U-shaped attention bias by reducing the attention weights between the initial token and tokens in the attention-dense areas, while increasing the attention weights between the initial token and tokens in the attention-sparse areas. Assume the input length is $n$. To achieve this, we propose a simple calibration technique comprising three steps:

1. Identify the attention-dense areas $Dense_{l}=\{i|TopA_{doc_{m} }^{n} > \sigma \cdot \frac{\sum TopA_{doc_{m} }^{n}}{M} , i\in doc_{m}\}$ for every layer $l \in L$. $doc_{m}$ represent the $m$-$th$ document and its corresponding region. $TopA_{doc_{m} }^{n}$ represent the quantity of attention scores within the top 30\% range in the $m$-$th$ document. Attention-sparse areas will be $Sparse_{l}=(0,n]$-$Dense_{l}$.

2. Scaling the attention weight between initial token and various tokens by setting $A_{i}^{0}=\alpha _{i}\times A_{i}^{0}$ for all $i\in (0,n]$. $\alpha _{i}$ is a hyperparameter that varies according to the location. When $i \in Dense_{l}$ we use $\alpha _{i}$ to eliminate the excessive attention weight, while $i \in Sparse_{l}$ using $\alpha _{i}$ to attract more attention.

3. Apply Step 2 to specific layers. This is because we find that scaling the initial token weight is not necessary at every layer. Specific details will be discussed in the following section.
\subsection{Results of Scaling Initial Token weight}
\paragraph{Evaluation Models}We investigat the impact of SIW on a wide range of open-source LLMs, including:
LLaMA-2-7B~\cite{llama}, Tulu-2-7B~\cite{ivison2023camels}, Vicuna-7B~\cite{vicuna} and Qwen1.5-7B~\cite{qwen}.

\paragraph{Evaluation Tasks}We validate the effectiveness of attention calibration with SIW from two perspectives. First, we examine the effectiveness of SIW in multi-document question answering and retrieval tasks. Specifically, we utilize the MDQA~\cite{lostinthemiddle} and KV-Retrieval~\cite{lostinthemiddle} datasets, which both contain ground truth at different positions in the prompt. The MDQA task includes 20 documents, comprising 2,655 data points. The KV retrieval task consists of 50 key-value pairs with an average length of about 2k tokens. All experiments use greedy decoding.

From the results in Table \ref{MDQA_1}, we can draw several conclusions: (1) SIW enhances the model’s ability to retrieve and understand key information from middle positions, alleviating the ``lost in the middle'' phenomenon. Compared to the baseline models, improvements ranging from 1.2\% to 9.2\% were observed across multiple open-source models for data where key information is located in the middle. (2) SIW provides an overall performance boost for the model in RAG and information retrieval tasks, achieving a maximum improvement of 3.6\% in MDQA and KV-Retrieval.
\begin{table*}[h!]
    \centering
    \small
    \renewcommand{\arraystretch}{1} 
    \begin{tabular}{cp{1.5cm}|p{0.6cm}p{0.6cm}p{0.6cm}p{0.6cm}p{0.6cm}p{0.6cm}|p{0.6cm}p{0.6cm}p{0.6cm}p{0.6cm}p{0.6cm}p{0.6cm}}
        \toprule
        \multicolumn{2}{c|}{\multirow{2}{*}{\textbf{Method}}} & \multicolumn{6}{c|}{\textbf{KV}} & \multicolumn{6}{c}{\textbf{MDQA}} \\
        & & 1 & 15 & 30 & 40 & 50 & avg & 1 & 5 & 10 & 15 & 20 & avg \\
        \midrule
        \multirow{2}{*}{Llama} & Base Model & \textbf{79.6} & \textbf{44.2} & 20.0 & 43.8 & 23.8 & 42.3 & 38.6 & 28.5 & 31.9 & 32.1 & 40.2 & 34.3 \\
         & SIW & 71.6 & 40.4 & \textbf{25.6} & \textbf{46.8} & \textbf{35.0} & \textbf{43.9} & \textbf{41.7} & \textbf{33.0} & \textbf{35.1} & \textbf{35.8} & \textbf{44.1} & \textbf{37.9} \\
         \midrule
        \multirow{2}{*}{Vicuna} & Base Model & \textbf{90.6} & 28.8 & 10.0 & 37.8 & \textbf{64.2} & 46.3 & \textbf{61.8} & 50.8 & 44.1 & 44.5 & 51.8 & 50.6 \\
         & SIW & 90.0 & \textbf{29.4} & \textbf{13.8} & \textbf{39.4} & 63.4 & \textbf{47.2} & 60.7 & \textbf{52.0} & \textbf{46.6} & \textbf{45.9} & \textbf{51.9} & \textbf{51.4} \\
         \midrule
         \multirow{2}{*}{Tulu} & Base Model & \textbf{99.6} & 59.0 & 38.4 & 60.2 & 32.0 & 57.2 & \textbf{45.9} & 30.4 & 30.1 & 32.4 & 46.0 & 37.0 \\
         & SIW & 96.4 & \textbf{59.4} & \textbf{47.6} & \textbf{61.2} & \textbf{32.8} & \textbf{59.5} & 45.0 & \textbf{32.5} & \textbf{32.1} & \textbf{32.7} & \textbf{47.0} & \textbf{37.9} \\
         \midrule
         \multirow{2}{*}{Qwen} & Base Model & 99.4 & \textbf{100} & 98.2 & 98.8 & \textbf{99.0} & 99.1 & \textbf{73.2} & 56.9 & 56.2 & \textbf{56.0} & \textbf{58.8} & 60.2 \\
         & SIW & \textbf{100} & \textbf{100} & \textbf{99.4} & \textbf{99.4} & 98.6 & \textbf{99.5} & 73.0 & \textbf{58.5} & \textbf{58.2} & 55.4 & 58.2 & \textbf{60.7} \\
        \bottomrule
    \end{tabular}
    \caption{Performance of SIW across different models on the MDQA and KV retrieval datasets.}
    \label{MDQA_1}
\end{table*}
\begin{figure}[t]
    \centering
    \includegraphics[width=0.4\textwidth]{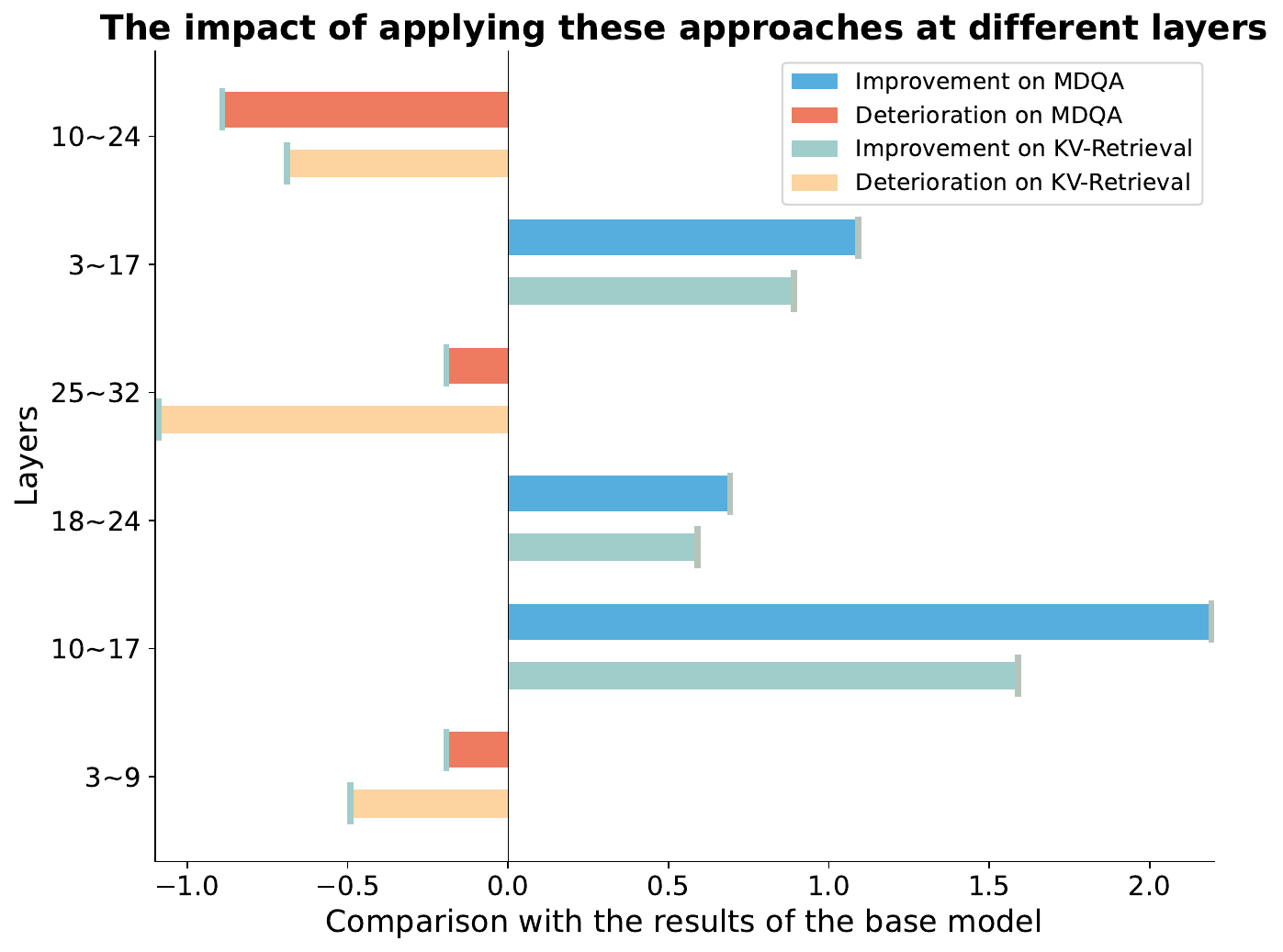}  
    \caption{The results of applying SIW across different layers.}
    \label{fig:sample_image}
\end{figure}
\begin{figure}[t]
    \centering
    \includegraphics[width=0.4\textwidth]{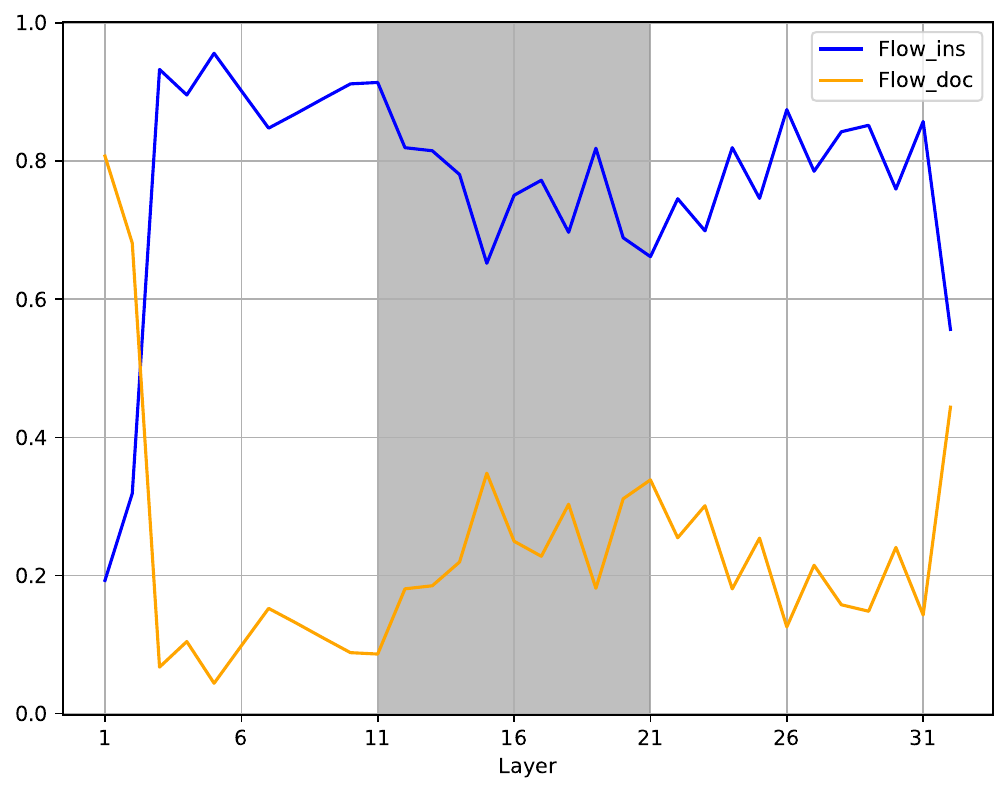}  
    \caption{The distribution of information flow between instructions and documents across different layers.}
    \label{fig:sc}
\end{figure}
\subsection{Results of SIW with Position Information Scaling}
As discussed earlier, the primary aim of SIW is to correct the bias in the first half of attention waves. Previous studies have extensively explored scaling position information to address bias in the latter half of attention waves. We also investigate whether using both approaches simultaneously to globally adjust attention waves could further improve the model’s performance on long-text RAG task. All experiments use greedy decoding.

We selecte three representative methods for scaling position information: (1) \textbf{SelfExtend}~(SE)~\cite{se} processes long- and short-range information separately by scaling positional encoding information and applying grouped and neighboring attention mechanisms. (2) \textbf{scale positional hidden states}~(Sphs)~\cite{caulsual_mask} scales position information by adjusting specific dimension within token representation vectors that encode positional information. We apply these two ideas simultaneously across multiple models and validated their effectiveness on MDQA\footnote{In order to ensure consistency with the experimental setups of methods such scale positional hidden states, we tested the first 500 samples of the MDQA dataset.} and KV-Retrieval dataset.

The results\footnote{To rigorously compare the experimental results before and after using SIW, we present the Sphs and SE results under the same experimental conditions.} in Table \ref{MDQA_2} show several key findings: (1) Building upon existing methods, SIW still improves the model's ability to retrieve and understand key information from middle positions, with a maximum improvement of 7.6\% for data where key information is located in the middle. (2) Results across multiple models demonstrate that SIW enhances various position information scaling methods, with a maximum improvement of 3.4\% in MDQA and KV-Retrieval. (3) Overall, SIW provides a performance boost across several position information scaling methods. This further confirms that simultaneously weakening initial saliency and position encoding bias more effectively reduces the U-shaped attention bias, thereby alleviating the ``lost in the middle'' phenomenon.
\subsection{Analysis of the Side Effects of SIW}
\begin{figure*}[t]
    \centering
    \subcaptionbox{Results of Llama-7B\label{longbench_1}}{\includegraphics[width=0.8\linewidth, height=0.2\linewidth]{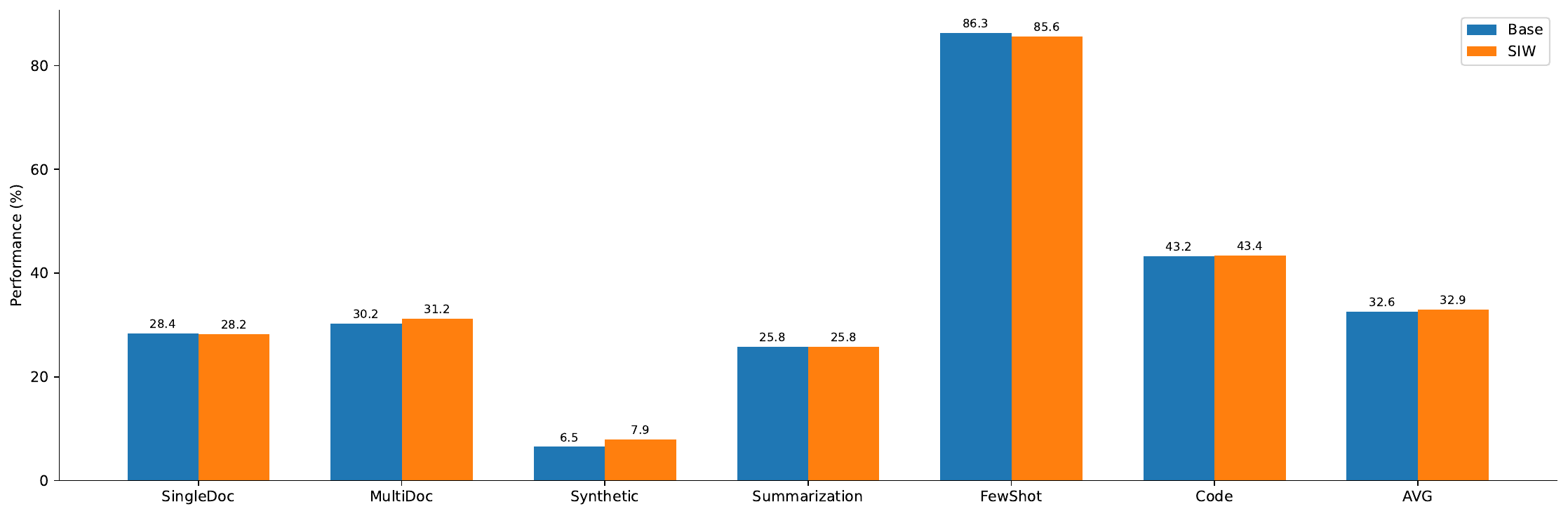}}
    \hspace{0.0001mm}
    \subcaptionbox{Results of Qwen-7B\label{longbench_2}}{\includegraphics[width=0.8\linewidth, height=0.2\linewidth]{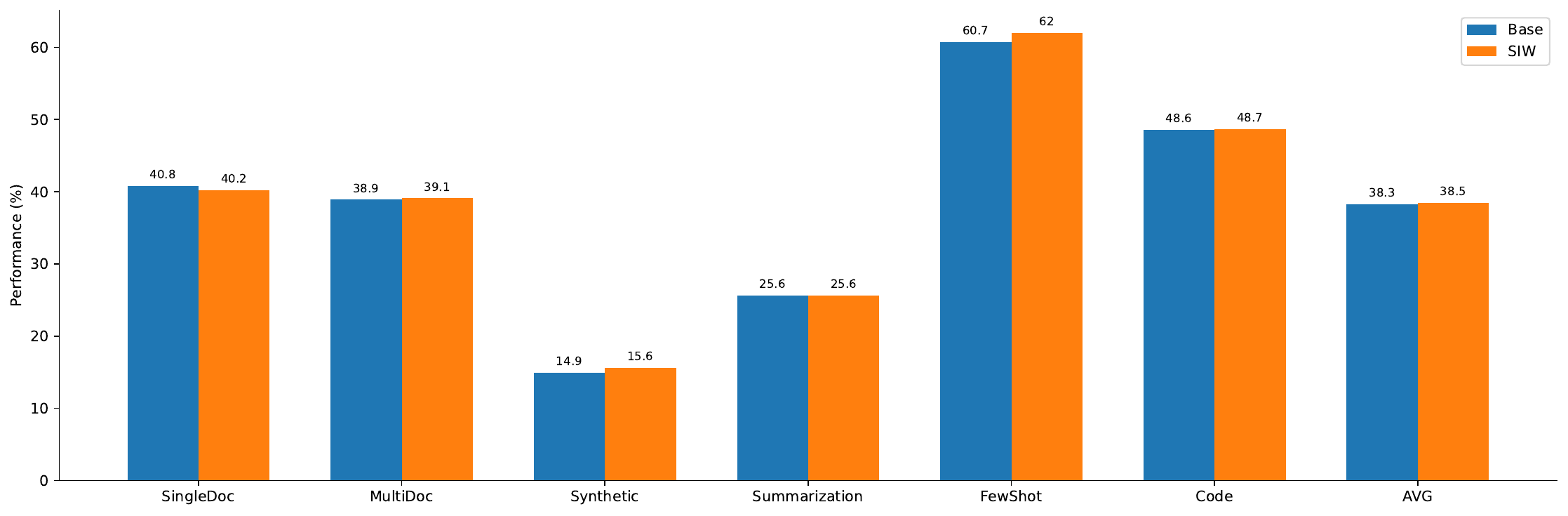}}
    \caption{Performance of the base model and the model after applying SIW on LongBench.}
    \label{longbench}
\end{figure*}

Our previous experiments have demonstrated that SIW significantly alleviates the ``lost in the middle'' phenomenon in tasks such as RAG and information extraction. Furthermore, we investigate the effects or potential side effects of SIW across a broader range of long-text tasks. To this end, we selecte LongBench~\cite{longbench}, a comprehensive benchmark for long-text tasks, which includes multi-document QA, single-document QA, summarization, few-shot learning, synthetic tasks, and code completion. The benchmark consists of 16 tasks with an average document length of 3.7k tokens. The results in Figure~\ref{longbench} indicate that SIW does not degrade performance on other long-text tasks. On the contrary, SIW overall enhances the model's general long-text performance, achieving an improvement of more than 0.2 in long-text tasks.
\begin{table*}[h!]
    \centering
    \renewcommand{\arraystretch}{0.82}
    \begin{tabular}{cp{2.2cm}|p{0.5cm}p{0.5cm}p{0.5cm}p{0.5cm}p{0.5cm}p{0.5cm}|p{0.5cm}p{0.5cm}p{0.5cm}p{0.5cm}p{0.5cm}p{0.5cm}}
        \toprule
        \multicolumn{2}{c|}{\multirow{2}{*}{\textbf{Method}}} & \multicolumn{6}{c|}{\textbf{KV}} & \multicolumn{6}{c}{\textbf{MDQA}} \\
        & & 1 & 15 & 30 & 40 & 50 & avg & 1 & 5 & 10 & 15 & 20 & avg \\
        \midrule
        \multirow{4}{*}{Llama} & SE & 96.6 & \textbf{56.8} & \textbf{76.9} & \textbf{90.2} & 80.2 & \textbf{80.1} & 42.6 & 29.6 & 29.4 & 38.6 & 39.2 & 35.9 \\
         & SE w SIW & \textbf{98.8} & 54.6 & 71.4 & 89.0 & \textbf{82.8} & 79.3 & \textbf{43.8} & \textbf{32.4} & \textbf{34.4} & \textbf{39.2} & \textbf{40.4} & \textbf{38.0} \\
         & Sphs & 60.4 & 44.0 & 45.2 & 72.0 & 87.4 & 61.8 & 33.6 & 34.0 & \textcolor{blue}{40.6} & 43.0 & 51.8 & 40.6 \\
         & Sphs w SIW & \textcolor{blue}{64.8} & \textcolor{blue}{45.8} & \textcolor{blue}{45.2} & \textcolor{blue}{73.2} & \textcolor{blue}{88.2} & \textcolor{blue}{63.4} & \textcolor{blue}{35.2} & \textcolor{blue}{34.6} & 40.2 & \textcolor{blue}{43.4} & \textcolor{blue}{53.8} & \textcolor{blue}{41.4} \\
         \midrule
        \multirow{4}{*}{Vicuna} & SE & \textbf{99.6} & \textbf{56.4} & 69.0 & 82.8 & 83.2 & 78.2 & 69.2 & 49.8 & 43.8 & 43.8 & \textbf{52.2} & 51.8 \\
         & SE w SIW & 99.6 & 56.0 & \textbf{78.8} & \textbf{84.6} & \textbf{83.4} & \textbf{80.5} & \textbf{69.6} & \textbf{51.4} & \textbf{47.8} & \textbf{44.6} & 51.4 & \textbf{53.0} \\
         & Sphs & 95.2 & 77.6 & 69.8 & 71.2 & 71.0 & 77.0 & \textcolor{blue}{57.2} & 47.6 & 40.8 & 42.8 & 53.8 & 48.4 \\
         & Sphs w SIW & \textcolor{blue}{95.4} & \textcolor{blue}{78.6} & \textcolor{blue}{75.6} & \textcolor{blue}{74.6} & \textcolor{blue}{72.0} & \textcolor{blue}{79.2} & 56.0 & \textcolor{blue}{49.0} & \textcolor{blue}{42.6} & \textcolor{blue}{44.0} & \textcolor{blue}{54.4} & \textcolor{blue}{49.2} \\
         \midrule
         \multirow{4}{*}{Tulu} & SE & - & - & - & - & - & - & 44.8 & 30.0 & 28.2 & 32.2 & \textbf{44.2} & 35.9 \\
         & SE w SIW & - & - & - & - & - & - & \textbf{45.0} & \textbf{31.2} & \textbf{30.2} & \textbf{32.4} & 43.8 & \textbf{36.5} \\
         & Sphs & 96.4 & \textcolor{blue}{59.2} & \textcolor{blue}{39.0} & 61.4 & 31.4 & 57.5 & 53.8 & 46.6 & 43.0 & \textcolor{blue}{47.0} & 55.4 & 49.2 \\
         & Sphs w SIW & \textcolor{blue}{97.2} & 59.0 & 38.4 & \textcolor{blue}{62.0} & \textcolor{blue}{40.6} & \textcolor{blue}{59.4} & \textcolor{blue}{55.4} & \textcolor{blue}{47.0} & \textcolor{blue}{44.0} & 46.2 & \textcolor{blue}{55.6} & \textcolor{blue}{49.6} \\
         \midrule
         \multirow{4}{*}{Qwen} & SE & 96.3 & 72.9 & 51.3 & \textbf{39.7} & 53.5 & 62.7 & \textbf{72.6} & 54.4 & 55.8 & 52.4 & \textbf{56.4} & 58.3  \\
         & SE w SIW & \textbf{96.5} & \textbf{81.1} & \textbf{58.9} & 39.5 & \textbf{54.7} & \textbf{66.1} & 72.0 & \textbf{57.2} & \textbf{56.6} & \textbf{52.4} & 55.2 & \textbf{58.7} \\
         & Sphs & \textcolor{blue}{93.4} & 98.0 & 93.8 & 98.2 & \textcolor{blue}{89.4} & 94.6 & \textcolor{blue}{65.6} & 54.0 & 51.4 & 54.4 & 58.2 & 56.7 \\
         & Sphs w SIW & 92.5 & \textcolor{blue}{98.8} & \textcolor{blue}{95.9} & \textcolor{blue}{98.7} & 87.8 & \textcolor{blue}{94.7} & 63.8 & \textcolor{blue}{56.0} & \textcolor{blue}{53.4} & \textcolor{blue}{55.4} & \textcolor{blue}{60.4} & \textcolor{blue}{57.8} \\
        \bottomrule
    \end{tabular}
    \caption{Performance of various methods across different models on the MDQA and KV retrieval datasets.}
    \label{MDQA_2}
\end{table*}
\begin{table}[h]
\renewcommand{\arraystretch}{1}
\centering
\begin{tabular}{@{}lcccc@{}}
\toprule
\textbf{Token name} & `<s>' & \texttt{`\textbackslash n'} & \texttt{`:'} & \texttt{`.'}  \\ \midrule
\textbf{Results}      & 34.3         & 33.5       & 34.1            & 32.5                   \\ \bottomrule
\end{tabular}
\caption{Performance of SIW on the MDQA dataset when different tokens serve as initial tokens.}
\label{toks}
\end{table}
\begin{table}[h!]
\centering
\renewcommand{\arraystretch}{1} 
\setlength{\tabcolsep}{4pt} 
\begin{tabular}{@{}p{1.5cm}cccccc@{}}
\toprule
\textbf{Position} & 0 & 499 & 999 & 1499 & 1999 \\ \midrule
\textbf{Attention} & 0.33 & 0.24 & 0.23 & 0.21 & 0.16 \\ \bottomrule
\end{tabular}
\caption{Attention weights assigned to `<s>' at different positions in the prompt.}
\label{atts}
\end{table}
\section{Analysis}
\subsection{Applying SIW in the Intermediate Layers Yields the Best}
The U-shaped attention bias is prevalent across most layers of the model. However, applying attention balancing to all these layers does not yield the best results. Experimental findings in Figure~\ref{fig:sample_image} indicate that using SIW for attention balancing in the middle layers alone achieves the desired effect.

This phenomenon may be related to the task-specific functions of different layers in LLMs. Previous research~\cite{layerskip1,skean2025layer,skean2024does} has found that LLMs are able to generate the final answer in the middle layers, while the later layers focus on verifying the accuracy of output. The intermediate layers of the model resemble \textbf{cognitive-intensive} layers. These layers place greater focus on the given textual information in order to generate a response. 

To validate this, we visualize the flow of information ~\cite{wang2023label} of attention for each layer, as shown in Figure \ref{fig:sc}. The task involves the first 500 data points in MDQA. We divide the input into two parts: Instruction and Documents, to analyze the flow of information. The Instruction refers to the part of the input that describes the task and output specifications to the model. The Documents concentrated in the middle of the input contain the information that the model needs to filter and utilize. Details of the flow of information computation are provided in the Appendix~\ref{sec:flow}. From Figure \ref{fig:sc}, it is evident that the information inflow from the Documents exhibits distinct peaks in the intermediate layers. This observation further corroborates that the intermediate layers are indeed the \textbf{cognitive-intensive} layers of the model, where it actively retrieves and processes information. 

Therefore, the intermediate layers function more like cognitive-intensive layers, playing a crucial role in reasoning. In these regions, encouraging the model to focus more on previously attention-scarce areas can significantly enhance its performance on long-text tasks.
\subsection{Impact of Different Tokens Acting as Initial Token on SIW}
Initial token is typically represented by the start token `<s>'. Existing research~\cite{streamllm,scale_AS} has also suggested that other tokens, such as punctuation marks (`\textbackslash n', `.', `:') or format control symbols, can serve as initial token. We also conducted an experiment to investigate the effect of different tokens acting as initial token on SIW. 

The experimental data was drawn from the first 500 samples of the MDQA dataset. As shown in Table \ref{toks}, the results of Llama-7B demonstrate that SIW improves the model's long-text capabilities regardless of the token serving as the SIW. But `<s>' yields the most significant effect. In further investigation, we examined the attention allocated to the token `<s>' at various positions within a random 2,000-token sequence. The results, shown in Table \ref{atts}, reveal that `<s>' consistently receives high attention, even when it is not placed at the beginning. No similar phenomenon was observed for other tokens. This indicates that `<s>' inherently functions as a \textbf{``black hole''} in attention computations. When positioned at the start, this effect is amplified, and tokens with similar vector representations to `<s>' also attract more attention. Consequently, SIW exhibits greater effectiveness in this scenario.
\section{Related Works}
\textbf{Struggles of long-context utilization in LLMs and U-shaped attention bias.} Although LLMs excel in a wide range of tasks~\cite{Radford_Narasimhan_Salimans_Sutskever,achiam2023gpt,llama,team2024gemini,liu2024deepseek}, their performance is still constrained by the limitations of the training corpus. In this context, RAG techniques remain crucial~\cite{gao2023retrieval}. LLMs require external information to address time-sensitive queries~\cite{fan2024survey} or provide more accurate responses in specialized fields such as healthcare~\cite{wang2024knowledge}. However, as the amount of supplemental information increases, studies have shown that LLMs struggle with long-context~\cite{shi2023large,li-etal-2024-loogle,li2024longcontext}. This phenomenon, known as ``lost in the middle'', refers to the model's tendency to focus on the beginning and end of a long context, neglecting information from the middle~\cite{lostinthemiddle}. This phenomenon has been found to be closely related to the U-shaped attention bias~\cite{middle}. However, previous studies have primarily focused on the impact of position encoding bias on attention bias, which does not fully explain the emergence of the U-shaped pattern.

\textbf{Improving long-context utilization in LLMs.} Research on position encoding bias~\cite{bucket,mspoe} has led to the development of various methods for enhancing the processing of long texts by leveraging scale-based positional information~\cite{chen2023extending,peng2023yarn}. These methods primarily involve either scaling position encodings~\cite{se,mspoe} or directly scaling the vectors that represent positions~\cite{caulsual_mask}. Besides, there are approaches such as the SFT-based method, which constructs data with more diverse key information distributions to train the model~\cite{he2024never,an2024make}. Another approach involves using special tokens to store compressed textual information~\cite{zhang2024soaring,gecontext}. Furthermore, some methods utilize external modules to reorder or compress information within the prompt, thereby enhancing the model's ability to handle long texts~\cite{han-etal-2024-lm,peysakhovich2023attention,jiang-etal-2024-longllmlingua}.

\section{Conclusion}
We have provided a comprehensive analysis of the factors contributing to the formation of U-shaped attention bias. Specifically, we introduced the concept of initial saliency and demonstrated its impact on attention distribution. Furthermore, our investigation revealed that balancing attention with scaling the initial token weight effectively mitigates the influence of initial saliency. Importantly, this approach can complement existing methods, collectively enhancing the model's capacity for handling long-text processing.

\section*{Limitations}
This paper introduces the concept of initial saliency, providing a more comprehensive explanation for the causes of U-shaped attention bias. Building on this, it enhances the model’s long-text performance by scaling the initial token weight and improves the effectiveness of existing position information scaling methods. However, this study does not conclusively rule out the possibility that factors beyond initial saliency and position encoding bias contribute to U-shaped attention bias. Additionally, the analysis of the causes of initial saliency remains incomplete. In our experiments, we also found that U-shaped attention may have an inherent rationale. In the real world, critical information often appears at the beginning and end of a sequence. Through extensive training, this attention distribution pattern may indeed be the most optimal.

\section*{Ethical Statement}
Our study utilizes publicly available datasets, ensuring that no personally identifiable information (PII) or sensitive user data is included. The primary goal of our work is academic research and the advancement of NLP technology. Our model is not intended for high-stakes applications. To promote transparency and reproducibility, we have submitted our code through the submission system, and it will be made publicly available on GitHub soon.
\bibliography{custom}

@article{scale_AS,
  title={Unveiling and Harnessing Hidden Attention Sinks: Enhancing Large Language Models without Training through Attention Calibration},
  author={Yu, Zhongzhi and Wang, Zheng and Fu, Yonggan and Shi, Huihong and Shaikh, Khalid and Lin, Yingyan Celine},
  booktitle={International Conference on Machine Learning},
  year={2024}
}

@article{caulsual_mask,
  title={Mitigate Position Bias in Large Language Models via Scaling a Single Dimension},
  author={Yu, Yijiong and Jiang, Huiqiang and Luo, Xufang and Wu, Qianhui and Lin, Chin-Yew and Li, Dongsheng and Yang, Yuqing and Huang, Yongfeng and Qiu, Lili},
  journal={arXiv preprint arXiv:2406.02536},
  year={2024}
}

@article{mspoe,
  title={Found in the Middle: How Language Models Use Long Contexts Better via Plug-and-Play Positional Encoding},
  author={Zhang, Zhenyu and Chen, Runjin and Liu, Shiwei and Yao, Zhewei and Ruwase, Olatunji and Chen, Beidi and Wu, Xiaoxia and Wang, Zhangyang},
  journal={arXiv preprint arXiv:2403.04797},
  year={2024}
}

@inproceedings{middle,
  title={Found in the middle: Calibrating Positional Attention Bias Improves Long Context Utilization},
  author={Hsieh, Cheng-Yu and Chuang, Yung-Sung and Li, Chun-Liang and Wang, Zifeng and Le, Long and Kumar, Abhishek and Glass, James and Ratner, Alexander and Lee, Chen-Yu and Krishna, Ranjay and others},
  booktitle={Findings of the Association for Computational Linguistics ACL 2024},
  pages={14982--14995},
  year={2024}
}

@article{lostinthemiddle,
  title={Lost in the middle: How language models use long contexts},
  author={Liu, Nelson F and Lin, Kevin and Hewitt, John and Paranjape, Ashwin and Bevilacqua, Michele and Petroni, Fabio and Liang, Percy},
  journal={Transactions of the Association for Computational Linguistics},
  volume={12},
  pages={157--173},
  year={2024},
  publisher={MIT Press One Broadway, 12th Floor, Cambridge, Massachusetts 02142, USA~…}
}

@article{longbench,
  title={Longbench: A bilingual, multitask benchmark for long context understanding},
  author={Bai, Yushi and Lv, Xin and Zhang, Jiajie and Lyu, Hongchang and Tang, Jiankai and Huang, Zhidian and Du, Zhengxiao and Liu, Xiao and Zeng, Aohan and Hou, Lei and others},
  journal={arXiv preprint arXiv:2308.14508},
  year={2023}
}

@article{llama,  
 title={Llama 2: Open Foundation and Fine-Tuned Chat Models}, 
 author={Touvron, Hugo and Martin, Louis and Stone, Kevin and Albert, Peter and Almahairi, Amjad and Babaei, Yasmine and Bashlykov, Nikolay and Batra, Soumya and Bhargava, Prajjwal and Bhosale, Shruti and Bikel, Dan and Blecher, Lukas and Ferrer, CristianCanton and Chen, Moya and Cucurull, Guillem and Esiobu, David and Fernandes, Jude and Fu, Jeremy and Fu, Wenyin and Fuller, Brian and Gao, Cynthia and Goswami, Vedanuj and Goyal, Naman and Hartshorn, Anthony and Hosseini, Saghar and Hou, Rui and Inan, Hakan and Kardas, Marcin and Kerkez, Viktor and Khabsa, Madian and Kloumann, Isabel and Korenev, Artem and Koura, Singh and Lachaux, Marie-Anne and Lavril, Thibaut and Lee, Jenya and Liskovich, Diana and Lu, Yinghai and Mao, Yuning and Martinet, Xavier and Mihaylov, Todor and Mishra, Pushkar and Molybog, Igor and Nie, Yixin and Poulton, Andrew and Reizenstein, Jeremy and Rungta, Rashi and Saladi, Kalyan and Schelten, Alan and Silva, Ruan and Michael, Eric and Ranjan, Smith and Xiaoqing, Subramanian and Tan, Ellen and Tang, Binh and Taylor, Ross and Williams, Adina and Kuan, JianXiang and Xu, Puxin and Yan, Zheng and Zarov, Iliyan and Zhang, Yuchen and Fan, Angela and Kambadur, Melanie and Narang, Sharan and Rodriguez, Aurelien and Stojnic, Robert and Edunov, Sergey and Scialom, Thomas and Genai, Meta}, 
 language={en-US},
 year={2023}
 }

@article{vicuna,  
 title={Vicuna: An Open-Source Chatbot Impressing GPT-4 with 90% ChatGPT Quality}, 
 author={Berkeley, UC and Cmu, Stanford and San, UC}, 
 language={en-US},
year={2023}
 }

@article{qwen,
  title={Qwen technical report},
  author={Bai, Jinze and Bai, Shuai and Chu, Yunfei and Cui, Zeyu and Dang, Kai and Deng, Xiaodong and Fan, Yang and Ge, Wenbin and Han, Yu and Huang, Fei and others},
  journal={arXiv preprint arXiv:2309.16609},
  year={2023}
}

@article{layerskip1,
  title={Layer skip: Enabling early exit inference and self-speculative decoding},
  author={Elhoushi, Mostafa and Shrivastava, Akshat and Liskovich, Diana and Hosmer, Basil and Wasti, Bram and Lai, Liangzhen and Mahmoud, Anas and Acun, Bilge and Agarwal, Saurabh and Roman, Ahmed and others},
  journal={arXiv preprint arXiv:2404.16710},
  year={2024}
}

@inproceedings{se,
  title={LLM Maybe LongLM: SelfExtend LLM Context Window Without Tuning},
  author={Jin, Hongye and Han, Xiaotian and Yang, Jingfeng and Jiang, Zhimeng and Liu, Zirui and Chang, Chia-Yuan and Chen, Huiyuan and Hu, Xia},
  booktitle={Forty-first International Conference on Machine Learning},
year={2024}
}

@article{snapkv,
  title={Snapkv: Llm knows what you are looking for before generation},
  author={Li, Yuhong and Huang, Yingbing and Yang, Bowen and Venkitesh, Bharat and Locatelli, Acyr and Ye, Hanchen and Cai, Tianle and Lewis, Patrick and Chen, Deming},
  journal={arXiv preprint arXiv:2404.14469},
  year={2024}
}

@article{streamllm,
  title={Efficient streaming language models with attention sinks},
  author={Xiao, Guangxuan and Tian, Yuandong and Chen, Beidi and Han, Song and Lewis, Mike},
  journal={arXiv preprint arXiv:2309.17453},
  year={2023}
}

@article{Radford_Narasimhan_Salimans_Sutskever,  
 title={Improving Language Understanding by Generative Pre-Training}, 
 author={Radford, Alec and Narasimhan, Karthik and Salimans, Tim and Sutskever, Ilya}, 
 language={en-US},
year={2018}
 }

@article{achiam2023gpt,
  title={Gpt-4 technical report},
  author={Achiam, Josh and Adler, Steven and Agarwal, Sandhini and Ahmad, Lama and Akkaya, Ilge and Aleman, Florencia Leoni and Almeida, Diogo and Altenschmidt, Janko and Altman, Sam and Anadkat, Shyamal and others},
  journal={arXiv preprint arXiv:2303.08774},
  year={2023}
}

@article{team2024gemini,
  title={Gemini 1.5: Unlocking multimodal understanding across millions of tokens of context},
  author={Team, Gemini and Georgiev, Petko and Lei, Ving Ian and Burnell, Ryan and Bai, Libin and Gulati, Anmol and Tanzer, Garrett and Vincent, Damien and Pan, Zhufeng and Wang, Shibo and others},
  journal={arXiv preprint arXiv:2403.05530},
  year={2024}
}

@inproceedings{bucket,
  title={Fortify the Shortest Stave in Attention: Enhancing Context Awareness of
Large Language Models for Effective Tool-Use},
  author={Chen, Yuhan and Lv, Ang and Lin, Ting-En and Chen, Changyu and Wu, Yuchuan and Huang, Fei and Li, Yongbin and Yan, Rui},
  booktitle={Proceedings of the 62nd Annual Meeting of the Association for Computational Linguistics},
  pages={ 11160–11174},
  year={2024}
}

@article{Zhong_Liu_Xu_Zhu_Zeng_2022,  
 title={DialogLM: Pre-trained Model for Long Dialogue Understanding and Summarization}, 
 url={http://dx.doi.org/10.1609/aaai.v36i10.21432}, 
 DOI={10.1609/aaai.v36i10.21432}, 
 journal={Proceedings of the AAAI Conference on Artificial Intelligence}, 
 author={Zhong, Ming and Liu, Yang and Xu, Yichong and Zhu, Chenguang and Zeng, Michael}, 
 year={2022}, 
 month={Jul}, 
 pages={11765–11773}, 
 language={en-US} 
 }

@article{wang2024knowledge,
  title={Knowledge-tuning Large Language Models with Structured Medical Knowledge Bases for Trustworthy Response Generation in Chinese},
  author={Wang, Haochun and Zhao, Sendong and Qiang, Zewen and Li, Zijian and Liu, Chi and Xi, Nuwa and Du, Yanrui and Qin, Bing and Liu, Ting},
  journal={ACM Transactions on Knowledge Discovery from Data},
  year={2024},
  publisher={ACM New York, NY}
}

@inproceedings{fan2024survey,
  title={A survey on rag meeting llms: Towards retrieval-augmented large language models},
  author={Fan, Wenqi and Ding, Yujuan and Ning, Liangbo and Wang, Shijie and Li, Hengyun and Yin, Dawei and Chua, Tat-Seng and Li, Qing},
  booktitle={Proceedings of the 30th ACM SIGKDD Conference on Knowledge Discovery and Data Mining},
  pages={6491--6501},
  year={2024}
}

@article{gao2023retrieval,
  title={Retrieval-augmented generation for large language models: A survey},
  author={Gao, Yunfan and Xiong, Yun and Gao, Xinyu and Jia, Kangxiang and Pan, Jinliu and Bi, Yuxi and Dai, Yi and Sun, Jiawei and Wang, Haofen},
  journal={arXiv preprint arXiv:2312.10997},
  year={2023}
}

@misc{li2024longcontext,
      title={Long-context LLMs Struggle with Long In-context Learning}, 
      author={Tianle Li and Ge Zhang and Quy Duc Do and Xiang Yue and Wenhu Chen},
      year={2024},
      eprint={2404.02060},
      archivePrefix={arXiv},
      primaryClass={cs.CL}
}

@inproceedings{li-etal-2024-loogle,
    title = "{L}oo{GLE}: Can Long-Context Language Models Understand Long Contexts?",
    author = "Li, Jiaqi  and
      Wang, Mengmeng  and
      Zheng, Zilong  and
      Zhang, Muhan",
    editor = "Ku, Lun-Wei  and
      Martins, Andre  and
      Srikumar, Vivek",
    booktitle = "Proceedings of the 62nd Annual Meeting of the Association for Computational Linguistics (Volume 1: Long Papers)",
    month = aug,
    year = "2024",
    address = "Bangkok, Thailand",
    publisher = "Association for Computational Linguistics",
    url = "https://aclanthology.org/2024.acl-long.859",
    doi = "10.18653/v1/2024.acl-long.859",
    pages = "16304--16333",
}

@inproceedings{shi2023large,
  title={Large language models can be easily distracted by irrelevant context},
  author={Shi, Freda and Chen, Xinyun and Misra, Kanishka and Scales, Nathan and Dohan, David and Chi, Ed and Sch{\"a}rli, Nathanael and Zhou, Denny},
  booktitle={Proceedings of the 40th International Conference on Machine Learning},
  pages={31210--31227},
  year={2023}
}

@article{asch1946forming,
  title={Forming impressions of personality.},
  author={Asch, Solomon E},
  journal={The journal of abnormal and social psychology},
  volume={41},
  number={3},
  pages={258},
  year={1946},
  publisher={American Psychological Association}
}

@article{baddeley1993recency,
  title={The recency effect: Implicit learning with explicit retrieval?},
  author={Baddeley, Alan D and Hitch, Graham},
  journal={Memory \& Cognition},
  volume={21},
  pages={146--155},
  year={1993},
  publisher={Springer}
}

@inproceedings{wang2023label,
  title={Label Words are Anchors: An Information Flow Perspective for Understanding In-Context Learning},
  author={Wang, Lean and Li, Lei and Dai, Damai and Chen, Deli and Zhou, Hao and Meng, Fandong and Zhou, Jie and Sun, Xu},
  booktitle={Proceedings of the 2023 Conference on Empirical Methods in Natural Language Processing},
  pages={9840--9855},
  year={2023}
}

@article{chen2023extending,
  title={Extending context window of large language models via positional interpolation},
  author={Chen, Shouyuan and Wong, Sherman and Chen, Liangjian and Tian, Yuandong},
  journal={arXiv preprint arXiv:2306.15595},
  year={2023}
}

@article{peng2023yarn,
  title={Yarn: Efficient context window extension of large language models},
  author={Peng, Bowen and Quesnelle, Jeffrey and Fan, Honglu and Shippole, Enrico},
  journal={arXiv preprint arXiv:2309.00071},
  year={2023}
}

@article{liu2024deepseek,
  title={Deepseek-v3 technical report},
  author={Liu, Aixin and Feng, Bei and Xue, Bing and Wang, Bingxuan and Wu, Bochao and Lu, Chengda and Zhao, Chenggang and Deng, Chengqi and Zhang, Chenyu and Ruan, Chong and others},
  journal={arXiv preprint arXiv:2412.19437},
  year={2024}
}

@article{an2024make,
  title={Make Your LLM Fully Utilize the Context},
  author={An, Shengnan and Ma, Zexiong and Lin, Zeqi and Zheng, Nanning and Lou, Jian-Guang},
  journal={arXiv preprint arXiv:2404.16811},
  year={2024}
}

@inproceedings{he2024never,
  title={Never Lost in the Middle: Mastering Long-Context Question Answering with Position-Agnostic Decompositional Training},
  author={He, Junqing and Pan, Kunhao and Dong, Xiaoqun and Song, Zhuoyang and LiuYiBo, LiuYiBo and Qianguosun, Qianguosun and Liang, Yuxin and Wang, Hao and Zhang, Enming and Zhang, Jiaxing},
  booktitle={Proceedings of the 62nd Annual Meeting of the Association for Computational Linguistics (Volume 1: Long Papers)},
  pages={13628--13642},
  year={2024}
}

@article{zhang2024soaring,
  title={Soaring from 4k to 400k: Extending llm’s context with activation beacon},
  author={Zhang, Peitian and Liu, Zheng and Xiao, Shitao and Shao, Ninglu and Ye, Qiwei and Dou, Zhicheng},
  journal={arXiv preprint arXiv:2401.03462},
  volume={2},
  number={3},
  pages={5},
  year={2024}
}

@inproceedings{gecontext,
  title={In-context Autoencoder for Context Compression in a Large Language Model},
  author={Ge, Tao and Jing, Hu and Wang, Lei and Wang, Xun and Chen, Si-Qing and Wei, Furu},
  booktitle={The Twelfth International Conference on Learning Representations},
year={2024}
}

@inproceedings{han-etal-2024-lm,
    title = "{LM}-Infinite: Zero-Shot Extreme Length Generalization for Large Language Models",
    author = "Han, Chi  and
      Wang, Qifan  and
      Peng, Hao  and
      Xiong, Wenhan  and
      Chen, Yu  and
      Ji, Heng  and
      Wang, Sinong",
    editor = "Duh, Kevin  and
      Gomez, Helena  and
      Bethard, Steven",
    booktitle = "Proceedings of the 2024 Conference of the North American Chapter of the Association for Computational Linguistics: Human Language Technologies (Volume 1: Long Papers)",
    month = jun,
    year = "2024",
    address = "Mexico City, Mexico",
    publisher = "Association for Computational Linguistics",
    url = "https://aclanthology.org/2024.naacl-long.222/",
    doi = "10.18653/v1/2024.naacl-long.222",
    pages = "3991--4008",
    
}

@article{peysakhovich2023attention,
  title={Attention sorting combats recency bias in long context language models},
  author={Peysakhovich, Alexander and Lerer, Adam},
  journal={arXiv preprint arXiv:2310.01427},
  year={2023}
}

@inproceedings{jiang-etal-2024-longllmlingua,
    title = "{L}ong{LLML}ingua: Accelerating and Enhancing {LLM}s in Long Context Scenarios via Prompt Compression",
    author = "Jiang, Huiqiang  and
      Wu, Qianhui  and
      Luo, Xufang  and
      Li, Dongsheng  and
      Lin, Chin-Yew  and
      Yang, Yuqing  and
      Qiu, Lili",
    editor = "Ku, Lun-Wei  and
      Martins, Andre  and
      Srikumar, Vivek",
    booktitle = "Proceedings of the 62nd Annual Meeting of the Association for Computational Linguistics (Volume 1: Long Papers)",
    month = aug,
    year = "2024",
    address = "Bangkok, Thailand",
    publisher = "Association for Computational Linguistics",
    url = "https://aclanthology.org/2024.acl-long.91/",
    doi = "10.18653/v1/2024.acl-long.91",
    pages = "1658--1677",
}

@article{skean2025layer,
  title={Layer by Layer: Uncovering Hidden Representations in Language Models},
  author={Skean, Oscar and Arefin, Md Rifat and Zhao, Dan and Patel, Niket and Naghiyev, Jalal and LeCun, Yann and Shwartz-Ziv, Ravid},
  journal={arXiv preprint arXiv:2502.02013},
  year={2025}
}

@article{skean2024does,
  title={Does representation matter? exploring intermediate layers in large language models},
  author={Skean, Oscar and Arefin, Md Rifat and LeCun, Yann and Shwartz-Ziv, Ravid},
  journal={arXiv preprint arXiv:2412.09563},
  year={2024}
}

@article{ivison2023camels,
  title={Camels in a changing climate: Enhancing lm adaptation with tulu 2},
  author={Ivison, Hamish and Wang, Yizhong and Pyatkin, Valentina and Lambert, Nathan and Peters, Matthew and Dasigi, Pradeep and Jang, Joel and Wadden, David and Smith, Noah A and Beltagy, Iz and others},
  journal={arXiv preprint arXiv:2311.10702},
  year={2023}
}

\clearpage

\appendix
\begin{figure*}[t]
    \centering
    \subcaptionbox{Layer 11 attn-scores\label{1}}{\includegraphics[width = .24\linewidth]{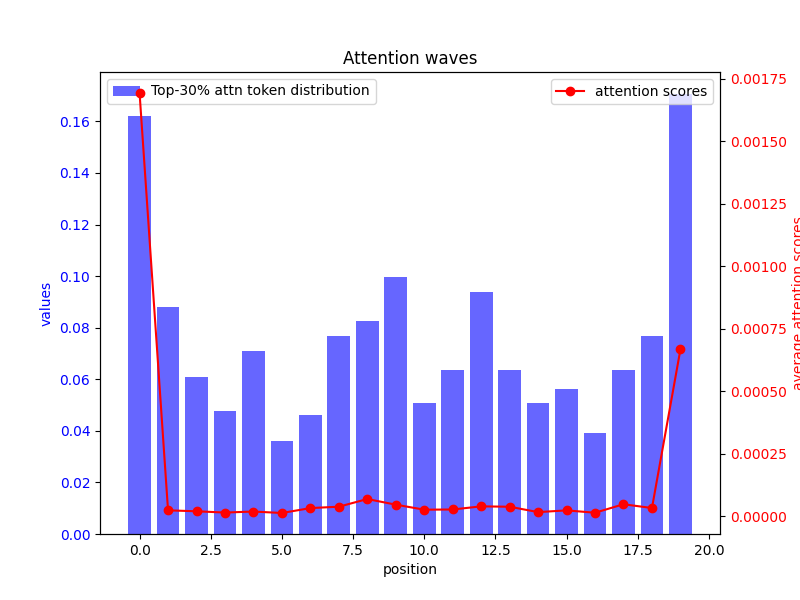}}
    \hspace{0.0001mm}
    \subcaptionbox{Layer 12 attn-scores\label{2}}{\includegraphics[width = .24\linewidth]{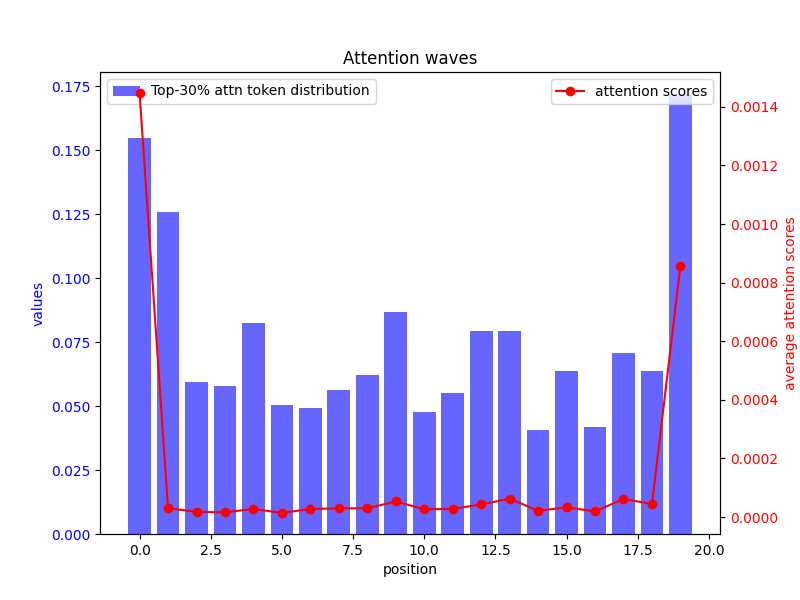}}
    \hspace{0.0001mm}
    \subcaptionbox{Layer 13 attn-scores\label{3}}
    {\includegraphics[width = .24\linewidth]{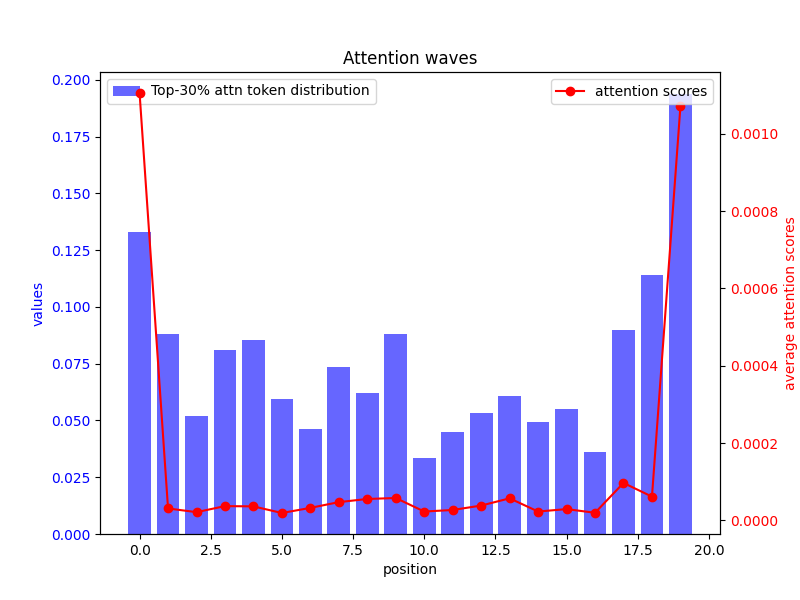}}
    \hspace{0.0001mm}
    \subcaptionbox{Layer 14 attn-scores\label{4}}
    {\includegraphics[width = .24\linewidth]{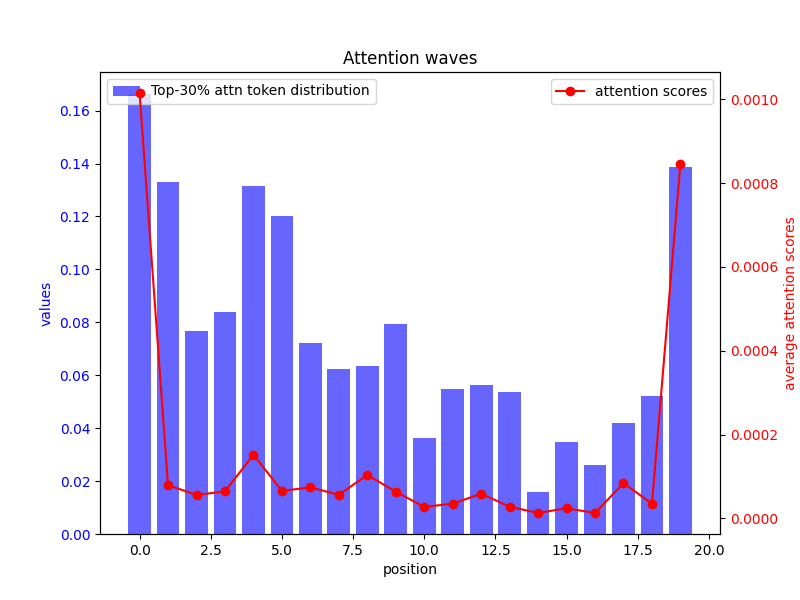}}
    \subcaptionbox{Layer 15 attn-scores\label{4}}
    {\includegraphics[width = .24\linewidth]{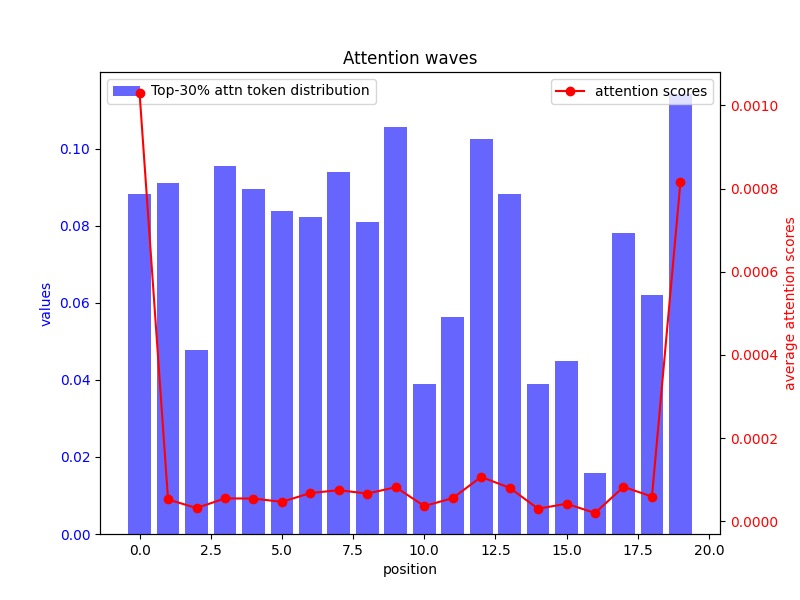}}
    \hspace{0.0001mm}
    \subcaptionbox{Layer 16 attn-scores\label{4}}
    {\includegraphics[width = .24\linewidth]{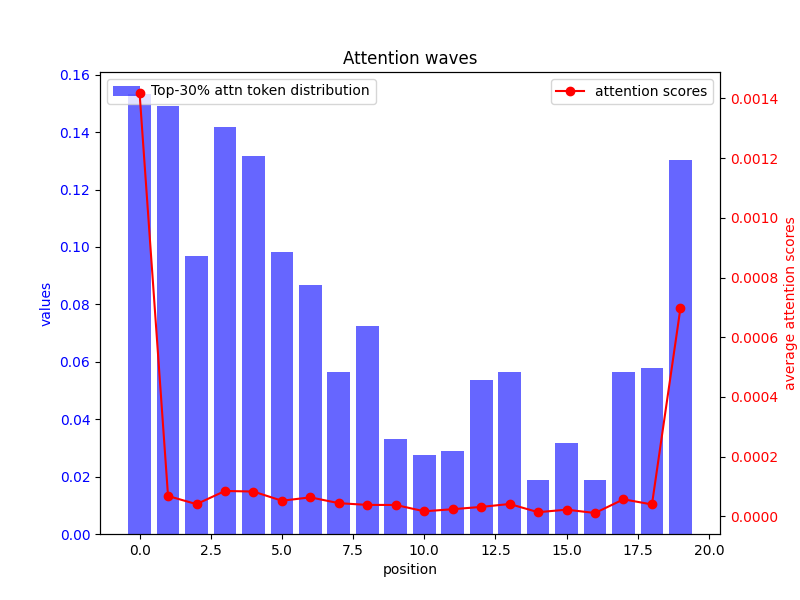}}
    \hspace{0.0001mm}
    \subcaptionbox{Layer 17 attn-scores\label{4}}
    {\includegraphics[width = .24\linewidth]{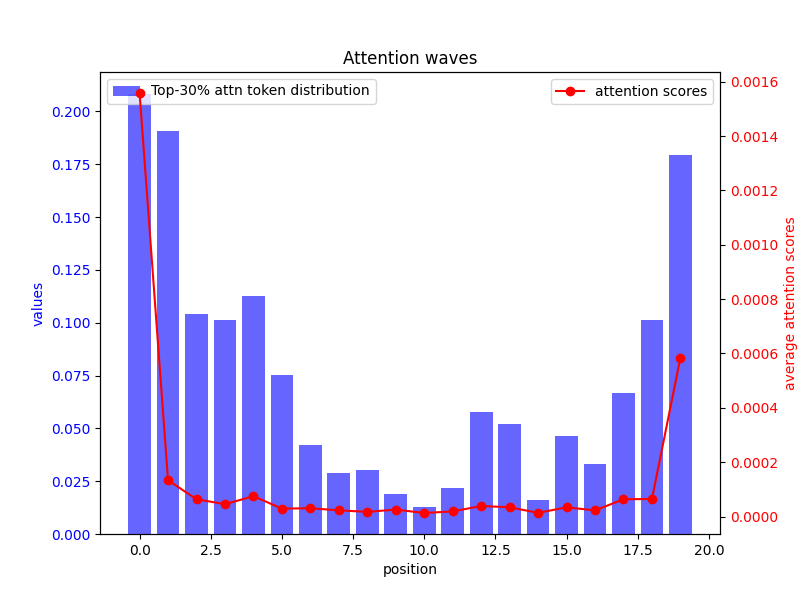}}
    \hspace{0.0001mm}
    \subcaptionbox{Layer 18 attn-scores\label{4}}
    {\includegraphics[width = .24\linewidth]{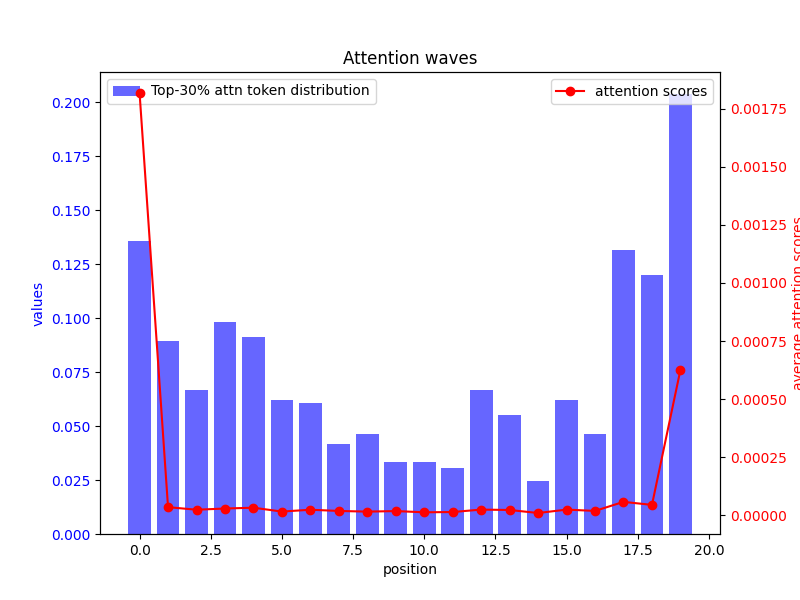}}
    \caption{ Visualization of the average attention in Llama2-7B-chat from layer 11 to layer 18.}
    \label{fig8}
\end{figure*}
\begin{figure*}
    \subcaptionbox{Layer 19 attn-scores\label{4}}
    {\includegraphics[width = .24\linewidth]{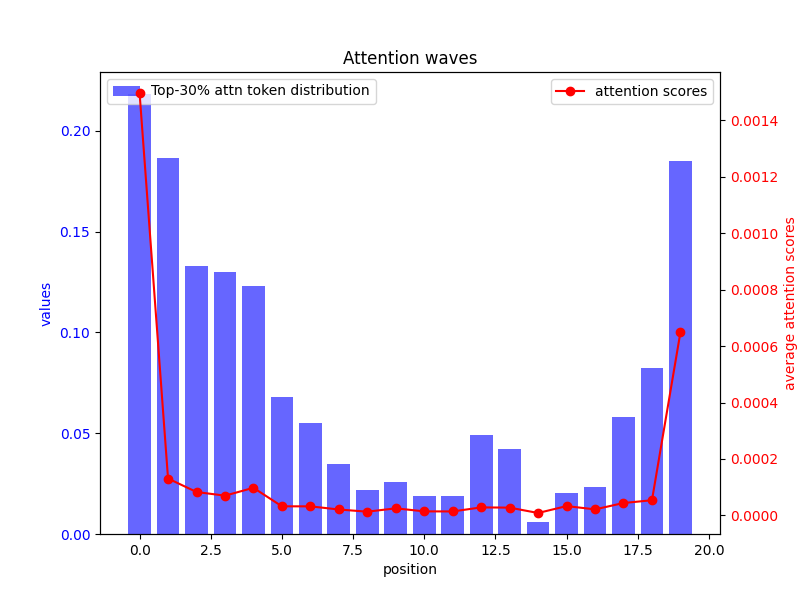}}
    \hspace{0.0001mm}
    \subcaptionbox{Layer 20 attn-scores\label{4}}
    {\includegraphics[width = .24\linewidth]{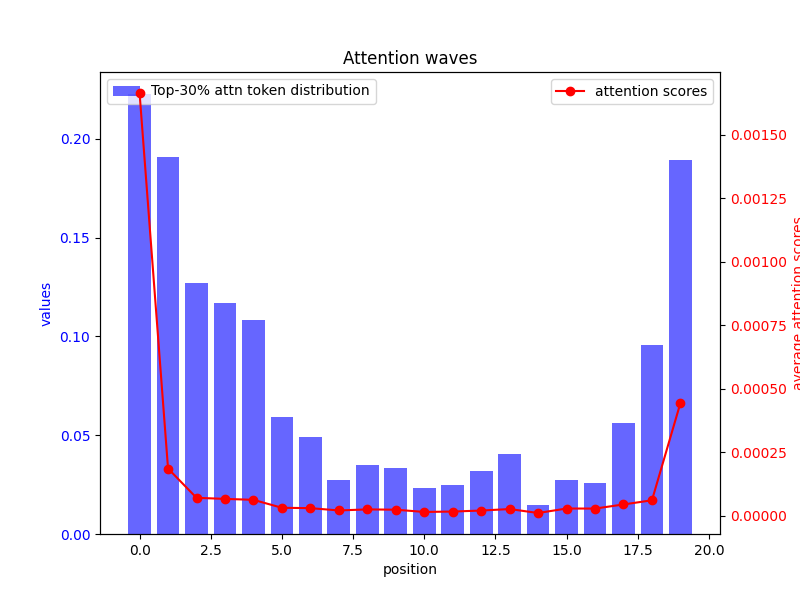}}
    \hspace{0.0001mm}
    \subcaptionbox{Layer 21 attn-scores\label{4}}
    {\includegraphics[width = .24\linewidth]{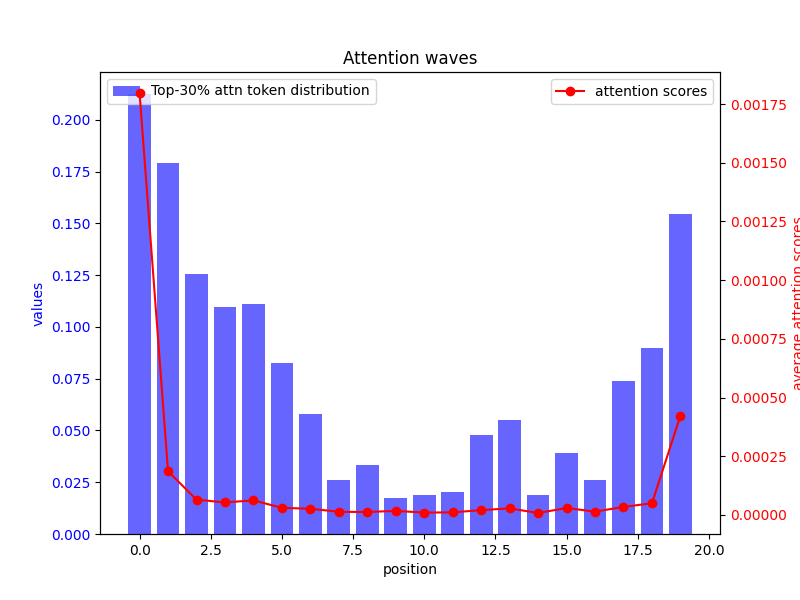}}
    \hspace{0.0001mm}
    \subcaptionbox{Layer 22 attn-scores\label{4}}
    {\includegraphics[width = .24\linewidth]{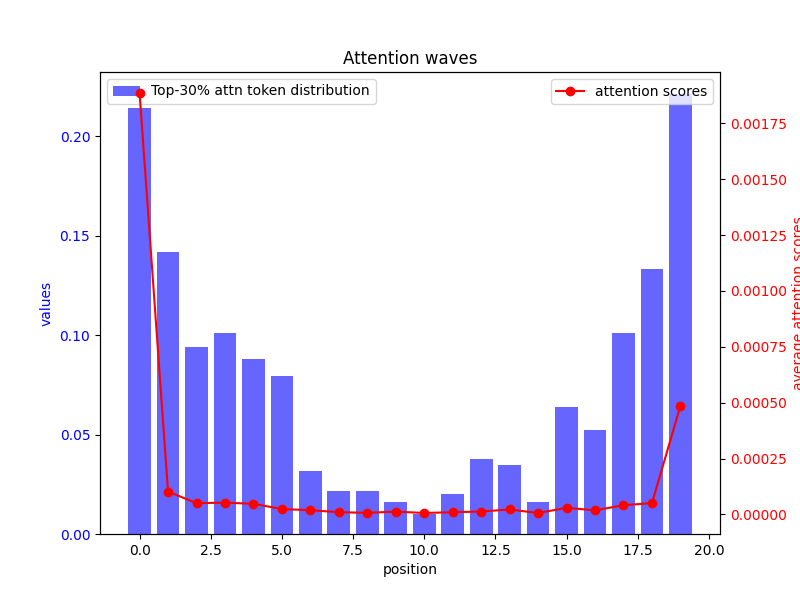}}
    \subcaptionbox{Layer 23 attn-scores\label{4}}
    {\includegraphics[width = .24\linewidth]{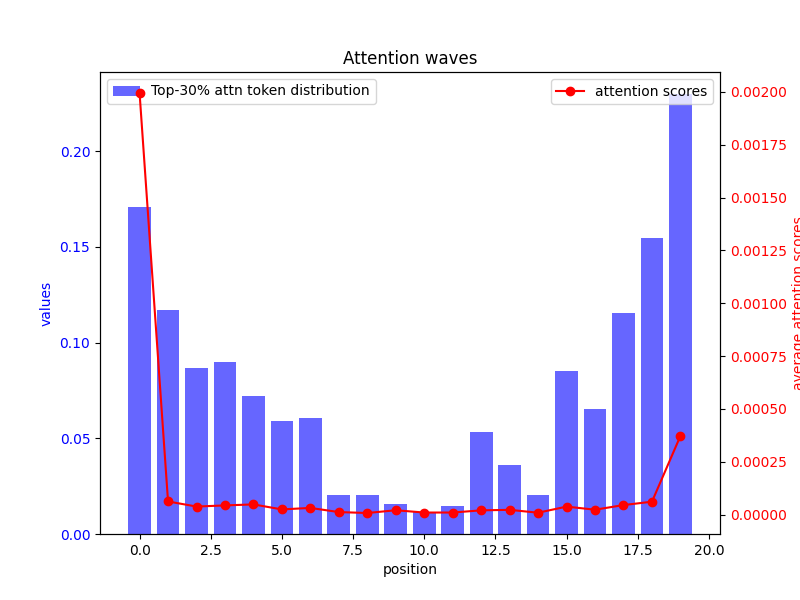}}
    \hspace{0.0001mm}
    \subcaptionbox{Layer 24 attn-scores\label{4}}
    {\includegraphics[width = .24\linewidth]{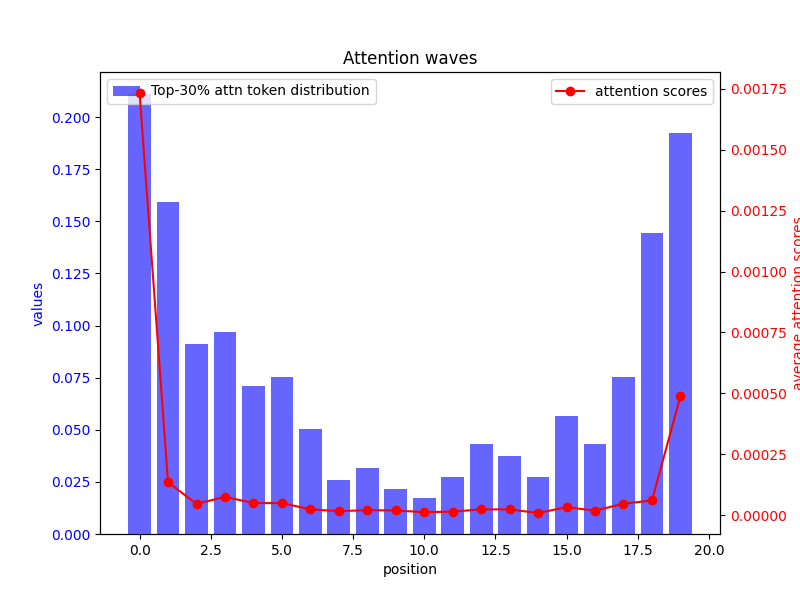}}
    \hspace{0.0001mm}
    \subcaptionbox{Layer 25 attn-scores\label{4}}
    {\includegraphics[width = .24\linewidth]{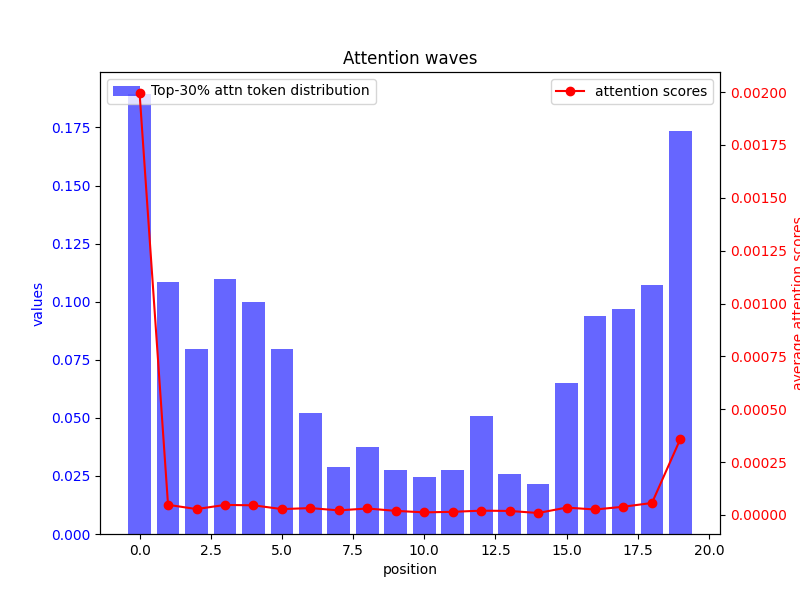}}
    \hspace{0.0001mm}
    \subcaptionbox{Layer 26 attn-scores\label{4}}
    {\includegraphics[width = .24\linewidth]{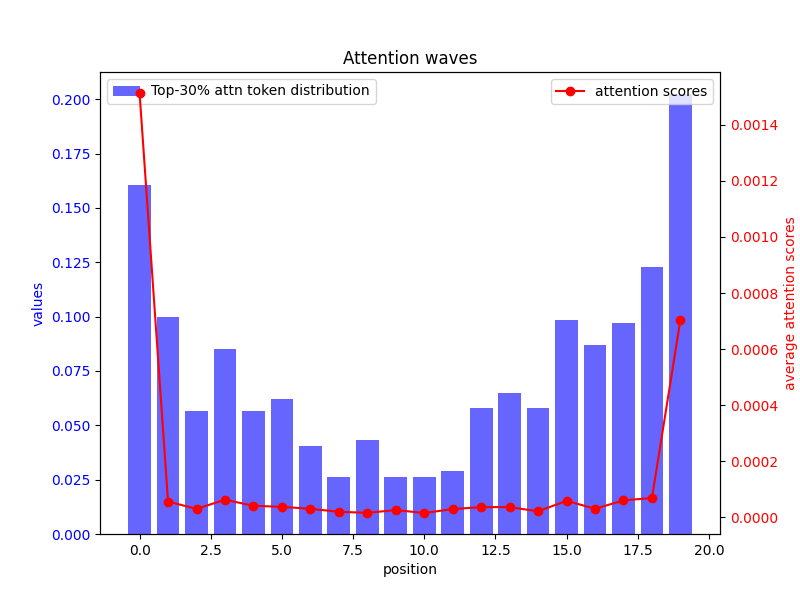}}
    \caption{ Visualization of the average attention in Llama2-7B-chat from layer 19 to layer 26.}
    \label{fig9}
\end{figure*}
\begin{figure}
    \subcaptionbox{Layer 27 attn-scores\label{4}}
    {\includegraphics[width = .48\linewidth]{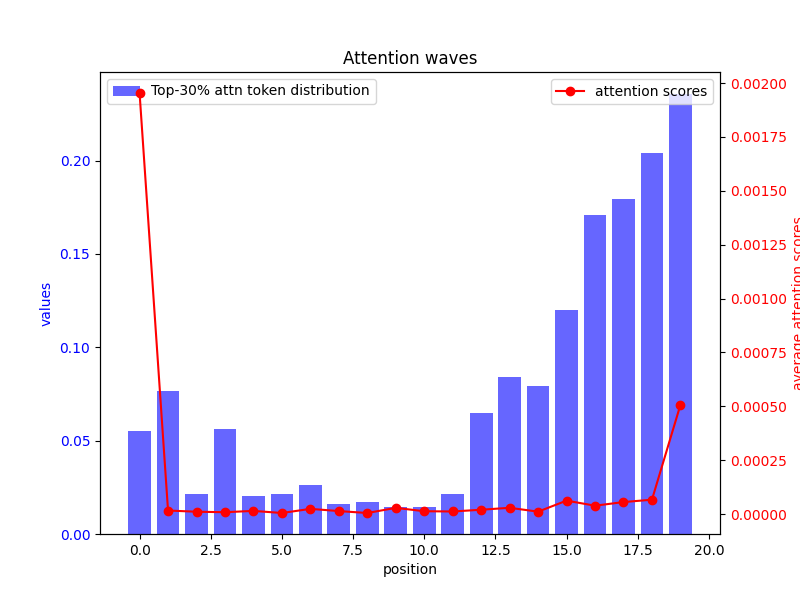}}
    \hspace{0.0001mm}
    \subcaptionbox{Layer 28 attn-scores\label{4}}
    {\includegraphics[width = .48\linewidth]{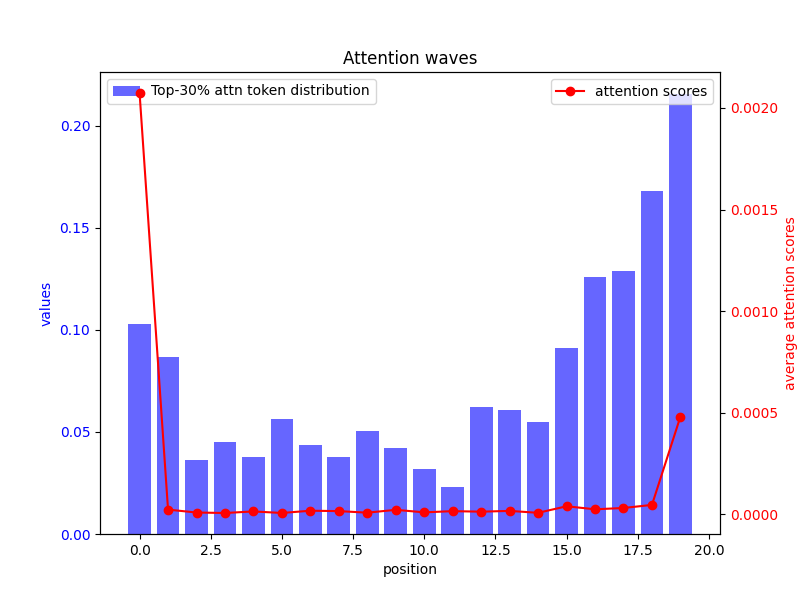}}
    \hspace{0.0001mm}
    \subcaptionbox{Layer 29 attn-scores\label{4}}
    {\includegraphics[width = .48\linewidth]{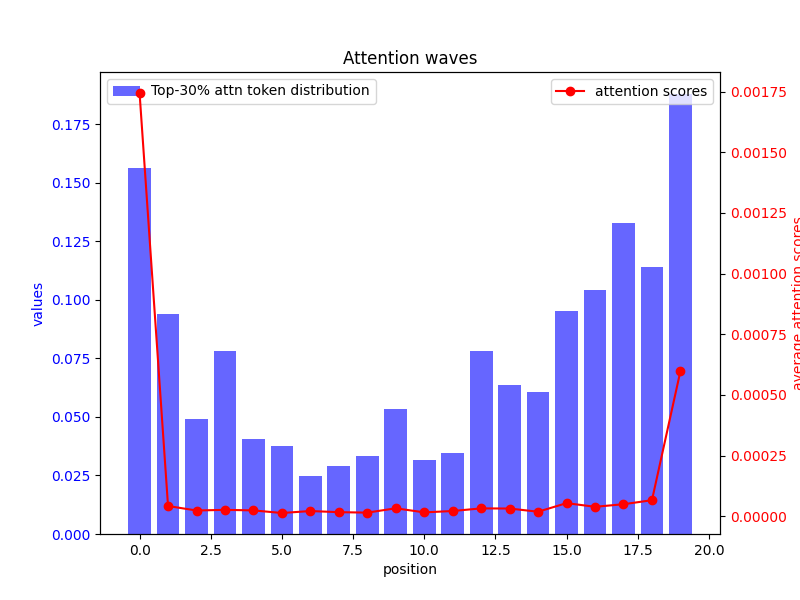}}
    \hspace{0.0001mm}
    \subcaptionbox{Layer 30 attn-scores\label{4}}
    {\includegraphics[width = .48\linewidth]{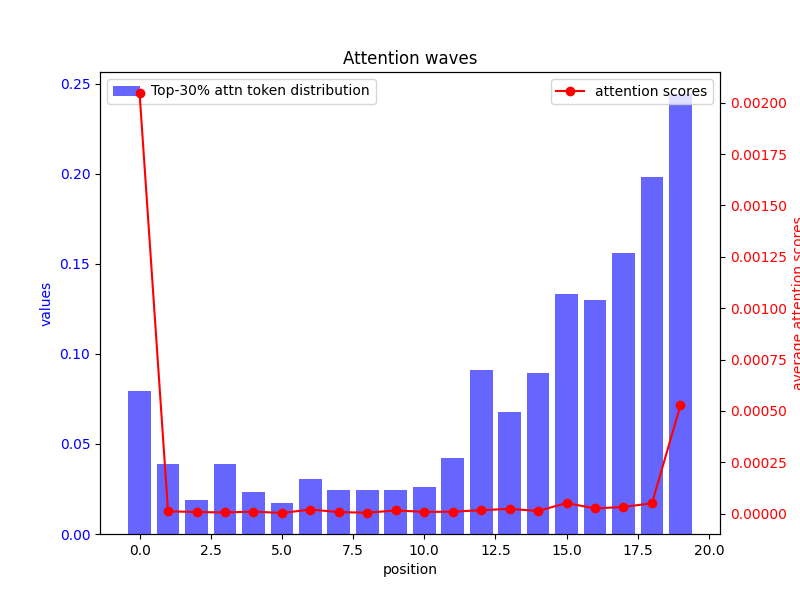}}
    \caption{ Visualization of the average attention in Llama2-7B-chat from layer 27 to layer 30.}
    \label{fig10}
\end{figure}
\section{The attention curves for Llama-7B}
\label{sec:barline}
Here, we present the attention curves for Llama-7B from layers 11 to 30 (Figure \ref{fig8}, Figure~\ref{fig9} and Figure~\ref{fig10}). The line chart represents the average attention weights of tokens across different documents, while the bar chart displays the distribution of the top 30\% attention-weighted tokens at various positions.
\section{Maintaining the Constraint That the Sum of Attention Weights Equals 1 Is Not Necessary for SIW}
Existing methods that use attention calibration often adhere to the constraint that the sum of attention weights equals 1. There are typically two approaches to achieve this: one involves normalizing the attention weights so that they sum to 1, while the other redistributes the excess attention to key tokens. We experimented with both approaches but observed a significant decline in performance. As noted in the side-effect analysis in previous sections, not enforcing the sum of attention weights to be 1 does not lead to significant adverse effects. This may be because the initial token itself usually does not carry semantic information. Therefore, it does not deeply impact the model's understanding or generation of semantics.
\begin{table}[h!]
\centering
\renewcommand{\arraystretch}{1.2} 
\setlength{\tabcolsep}{6pt} 
\begin{tabular}{@{}lcc@{}}
\toprule
\textbf{Method} & \textbf{Results} \\ \midrule
normalization    & 25.4          \\
focus on the vitalness             & 31.1             \\
do nothing & 34.3         \\ \bottomrule
\end{tabular}
\caption{The impact of maintaining the sum of attention equal to 1 in different methods on SIW.}
\end{table}
\section{Method of Information Flow Calculation}
\label{sec:flow}
We divide the input into two parts: Instruction and Documents, to analyze the flow of information. The Instruction refers to the part of the input that describes the task and output specifications to the model, and it is primarily concentrated at the beginning and end of the input. The Documents concentrated in the middle of the input contain the information that the model needs to filter and utilize.

The input $X$ is divided into $X^{doc}$ and $X^{ins}$, where $X^{doc}$ refers to the tokens in the document section and $X^{ins}$ refers to the tokens in the instruction section.The information flow, 
$Flow_{doc}$, represents the information flowing from the document tokens to the next token during next token prediction. 
\begin{equation}
    Flow_{doc}=\frac{\sum_{d=0}^{l} A_{d}^{l},~~tok_{d}\in X^{doc}  }{ \sum_{i=0}^{l} A_{i}^{l}  } 
\end{equation}
Similarly, the information flow, $Flow_{ins}$, represents the information flowing from the instruction tokens to the next token during next token prediction.
\begin{equation}
    Flow_{ins}=\frac{\sum_{c=0}^{l} A_{c}^{l},~~tok_{c}\in X^{doc}  }{ \sum_{i=0}^{l} A_{i}^{l}  } 
\end{equation}

\end{document}